\documentclass{article}

\usepackage{microtype}
\usepackage{graphicx}
\usepackage{booktabs} 

\usepackage{hyperref}

\usepackage[accepted]{icml2024}

\usepackage{amsmath}
\usepackage{amssymb}
\usepackage{mathtools}
\usepackage{amsthm}
\usepackage{multirow}
\usepackage{comment}
\usepackage{color, colortbl}
\usepackage{enumitem}
\usepackage{scalerel}
\usepackage{lipsum}
\usepackage{duckuments}
\usepackage{tcolorbox}
\usepackage{subfig}
\tcbuselibrary{breakable}

\usepackage[capitalize,noabbrev]{cleveref}

\theoremstyle{plain}

\theoremstyle{definition}

\theoremstyle{remark}

\newcolumntype{P}[1]{>{\centering\arraybackslash}p{#1}}
\colorlet{shadecolor}{gray!30}
\definecolor{shamrockgreen}{rgb}{0.0, 0.62, 0.38}
\definecolor{urobilin}{rgb}{0.88, 0.68, 0.13}
\definecolor{sinopia}{rgb}{0.8, 0.25, 0.04}

\usepackage{xcolor,pifont}
\newcommand*\colourcheck[1]{%
  \expandafter\newcommand\csname #1check\endcsname{\textcolor{#1}{\ding{52}}}%
}
\colourcheck{shamrockgreen}
\colourcheck{urobilin}

\definecolor{commentcolor}{RGB}{110,154,155}   
\definecolor{defcolor}{RGB}{225,81,145}

\usepackage[textsize=tiny]{todonotes}

\icmltitlerunning{Investigating Pre-Training Objectives for Generalization in Vision-Based Reinforcement Learning}

\begin{document}

\twocolumn[
\icmltitle{Investigating Pre-Training Objectives for Generalization \\ in Vision-Based Reinforcement Learning}



\icmlsetsymbol{equal}{*}

\begin{icmlauthorlist}
\icmlauthor{Donghu Kim}{equal,sch}
\icmlauthor{Hojoon Lee}{equal,sch}
\icmlauthor{Kyungmin Lee}{equal,sch}
\icmlauthor{Dongyoon Hwang}{sch}
\icmlauthor{Jaegul Choo}{sch}
\end{icmlauthorlist}

\icmlaffiliation{sch}{KAIST}

\icmlcorrespondingauthor{Donghu Kim}{quagmire@kaist.ac.kr}

\icmlkeywords{Machine Learning, ICML}

\vskip 0.3in]



\printAffiliationsAndNotice{\icmlEqualContribution} 

\begin{abstract}
Recently, various pre-training methods have been introduced in vision-based Reinforcement Learning (RL). However, their generalization ability remains unclear due to evaluations being limited to in-distribution environments and non-unified experimental setups. To address this, we introduce the Atari Pre-training Benchmark (Atari-PB), which pre-trains a ResNet-50 model on 10 million transitions from 50 Atari games and evaluates it across diverse environment distributions.
Our experiments show that pre-training objectives focused on learning task-agnostic features (e.g., identifying objects and understanding temporal dynamics) enhance generalization across different environments. In contrast, objectives focused on learning task-specific knowledge (e.g., identifying agents and fitting reward functions) improve performance in environments similar to the pre-training dataset but not in varied ones. We publicize our codes, datasets, and model checkpoints at \url{https://github.com/dojeon-ai/Atari-PB}.
\end{abstract}

\section{Introduction}
\label{introduction}

The pretrain-then-finetune approach has become a standard practice in computer vision (CV) and natural language processing (NLP), renowned for its robust generalization across diverse tasks \cite{he2022mae, bubeck2023gpt4}. 
In vision-based Reinforcement Learning (RL), this approach is now gaining attraction as well \cite{levine2020offline_rl_survey, yang2023foundation_survey}, driven by the increasing availability of large-scale offline datasets \cite{grauman2022ego4d, padalkar2023rtx}.

\begin{figure}[t]
\begin{center}
\includegraphics[width=0.91\linewidth]{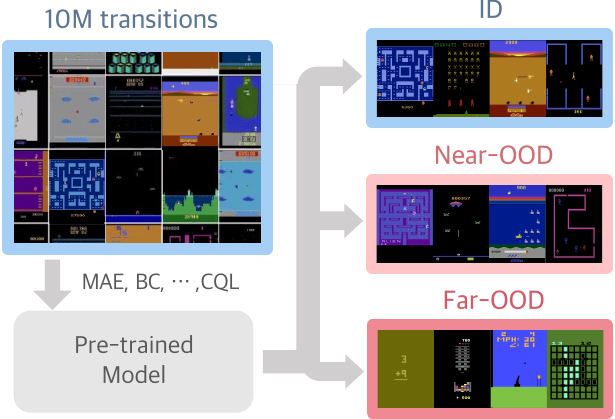}
\end{center}
\caption{\textbf{Overview of Atari-PB.} The ResNet-50-based model is pre-trained from 10M interactions with a given pre-training algorithm. The pre-trained model is then evaluated by fine-tuning to In-Distribution (ID), Near-Out-of-Distribution (Near-OOD), and Far-Out-of-Distribution (Far-OOD) environments.}
\label{fig:intro}
\end{figure}


In vision-based RL, various pre-training methods are designed to capture unique features from different data types. Image-based methods, for instance, focus on extracting spatial characteristics such as object sizes and shapes \cite{laskin2020curl, seo2023mwm}. In contrast, video-based approaches delve into the temporal dynamics of environments like objects' movement and direction \cite{schwarzer2020spr, nair2022r3m,gupta2023siammae}. 
Those learning from demonstrations prioritize extracting task-relevant features, distinguishing agents from irrelevant elements like backgrounds \cite{pomerleau1991bc, christiano2016idm, islam2022acro}. 
Finally, trajectory-based methods further concentrate on learning task-specific features by estimating the rewards associated with different states and actions \cite{kumar2020cql,chen2021dectransformer, fujimoto2019bcq}.

Despite their advancements, the generalization capabilities of these methods remain underexplored as evaluations are typically confined to environments akin to their pre-training datasets \cite{schwarzer2021sgi, lee2023simtpr}.
Several studies have probed the generalization ability of the agents by changing visual and task attributes, such as object colors \cite{hansen2021dmcgb}, shapes \cite{yuan2023rlvigen}, and physics \cite{taiga2022investigating}. Yet, these variations are relatively minor and may not sufficiently mirror the complexities found in real-world scenarios.
More recent research has begun to investigate the pre-trained models' generalization capabilities under larger distribution shifts \cite{nair2022r3m, xiao2022maskedvismotor, parisi2022pvr, majumdar2023cortexbench}. 
However, these studies use models pre-trained on different data sources and with varying architectures, complicating the understanding of how learning objectives affect generalization performance.

\begin{figure}[t]
\begin{center}
\includegraphics[width=1.0\linewidth]{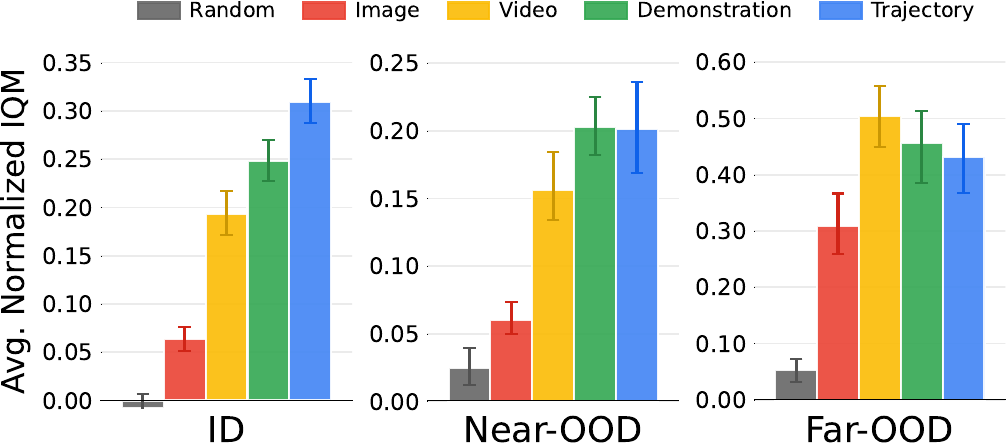}
\end{center}
\caption{\textbf{Results Overview.} The pre-training methods are evaluated by their performance after fine-tuning to environments in three groups: In-Distribution, Near-Out-of-Distribution, and Far-Out-of-Distribution. Here, we report the results of fine-tuning via behavior cloning (i.e., replicating expert behavior) and average the scores of each algorithm category for a comprehensive analysis.}
\label{fig:main_overview} 
\end{figure}

To address this gap, we introduce the Atari Pre-training Benchmark (Atari-PB), which investigates the generalization capability of various learning objectives using a unified dataset and architecture. Our benchmark begins by pre-training a ResNet-50 encoder \cite{he2016resnet} on a dataset containing 10 million interactions from 50 different Atari environments \cite{agarwal2020optimistic}. The pre-trained models are then fine-tuned across three groups of environments: In-Distribution (ID), Near-Out-of-Distribution (Near-OOD), and Far-Out-of-Distribution (Far-OOD). The ID group includes environments identical to those in the pre-training dataset. The Near-OOD group consists of similar tasks (e.g., shooting or tracking) but with different visual characteristics (e.g., object shape or speed). The Far-OOD group contains environments with entirely different tasks (e.g., solving math puzzles or color matching).

Our findings, illustrated in Figure \ref{fig:main_overview}, show that pre-training methods aimed at learning task-agnostic features, such as extracting spatial characteristics from images and temporal dynamics from videos, enhance generalization across various distribution shifts. 
In contrast, methods that learn task-specific knowledge, such as identifying agents from demonstrations and fitting reward functions from trajectories, enhance performance in the same environments but hinder generalization under distribution shifts.

\begin{tcolorbox}[breakable=true,
                  boxsep=0pt,
                  left=8pt,
                  right=9pt]
\textbf{Takeaways:}
\begin{itemize}[leftmargin=10pt, topsep=4pt]
\item Pre-training objectives that learn task-agnostic features, such as identifying objects from images and understanding temporal dynamics from videos, consistently improve generalization across various distribution shifts.
\item Pre-training objectives that learn task-specific features, such as focusing on agents from demonstrations and fitting the reward function from trajectories, improve performance in similar tasks but lose effectiveness as task distribution shifts increase.
\end{itemize}
\end{tcolorbox}

\section{Related Work}

\subsection{Evaluating Generalization in Vision-Based RL}

As Deep Reinforcement Learning has achieved notable success in specific games or tasks \cite{hafner2023dreamerv3, schwarzer2023bbf}, research interest is moving towards the agent's ability to generalize on environments with visual and/or task distribution shifts.

For visual generalization, researchers have focused on developing environments with controllable visual elements. 
This includes changing the design of objects and walls
\cite{lomonaco2020continual}, adding background noise \cite{hansen2021dmcgb, stone2021dcs}, and applying realistic disturbances such as changes in lighting and camera perspectives \cite{dosovitskiy2017carla, yuan2023rlvigen}.

Regarding task distribution shifts, numerous studies aimed to evaluate the agents on new tasks with similar dynamics but varying reward functions. For instance, \citet{farebrother2018generalization, taiga2022investigating} evaluated generalization abilities using different game modes and difficulty levels, while other works employed procedural generation \cite{cobbe2020procgen, zhu2020robosuite}.

Despite the importance of these studies, they have primarily considered a single type of distribution shift, often minor in nature. Thus, we propose evaluating agent performance in environments with novel tasks and visual shifts (Far-OOD), those with similar tasks and visual shifts (Near-OOD), and those with same tasks (ID). This broader coverage of distribution shifts enables a comprehensive assessment of agents' generalization abilities under visual and task shifts.

\subsection{Pre-training for Generalization in Vision-based RL}

The pretrain-then-finetune approach is a well-established framework in CV and NLP for improving generalization ability. 
Similarly for visual RL, the development of large-scale offline datasets \cite{grauman2022ego4d, padalkar2023rtx} have spurred investigations into the efficacy of pre-training methods for enhancing generalization capabilities.

Studies by \citet{parisi2022pvr} and \citet{hu2023policybench} have shown that the use of pre-trained models, such as ResNet \cite{he2016resnet} and ViT \cite{dosovitskiy2020vit}, enhances generalizability in various RL tasks. These models were pre-trained on ImageNet \cite{deng2009imagenet} and employed diverse training objectives, ranging from supervised \cite{radford2021clip} to self-supervised learning \cite{he2020mocov1, nair2022r3m}. Furthermore,  \citet{majumdar2023vc1} achieved improved generalization using a large, realistic dataset that closely resembles the ego-centric perspective of agents \cite{grauman2022ego4d}, to pre-train a large ViT with masked image modeling \cite{he2022mae}. 

Despite the advancements, these models were pre-trained with different datasets and architectures, making it challenging to identify the effectiveness of individual learning objectives for generalization. Our study addresses this gap by systematically evaluating the generalization capabilities of various pre-training objectives using a unified experimental setup, including an identical dataset, architecture, and downstream environments.

\section{Preliminaries}
\label{section:preliminary}

 \subsection{Reinforcement Learning}
In Reinforcement Learning (RL), we adopt a Partially Observable Markov Decision Process (POMDP) framework, which differs from standard MDP by limiting direct access to the true state $s_t$.
Instead, agents receive a partially observable image $o_t$, generated by an emission function: $o_t \sim q(\cdot|s_t)$. 
At each timestep, $t$, an agent observes an observation $o_t$ and chooses an action $a_t$ according to its policy $a_t \sim \pi(\cdot|o_t)$. The agent then receives the next observation $o_{t+1}$ and a reward $r_t$ from the environment. 
Here, the goal of the agent is to learn a policy that maximizes the expected cumulative reward, $\mathbb{E}[\sum^T_{t=1}r_t]$.

\subsection{Pre-training for Reinforcement Learning}
\label{subsection:method_class}

Drawing insights from the CV and NLP field, the pretrain-then-finetune approach has emerged as a compelling method to improve generalization, paving the way for models to effectively adapt to diverse tasks and environments.

When pre-training for RL, the model learns from the transitions made by humans or agents, using different objectives such as Offline RL \citep{kumar2020cql} and self-supervised loss \citep{laskin2020curl}.
These learning objectives are tightly related to the data type they leverage---image, video, demonstration, and trajectory---each introducing unique knowledge to the model.
Based on this, we classify the pre-training algorithms into the following four categories. Here, $N$ denotes the number of trajectories, $T$ denotes the length of the trajectories, and $NT$ denotes the total number of transitions in the dataset.

\begin{itemize}[itemsep=0pt, topsep=0pt]
\item \textbf{Image:} These methods learn the \textit{spatial characteristics} of individual states from a set of images, $\bigcup\limits_{i=1}^{NT}\{o_i\}$.
\item \textbf{Video:} These methods learn the \textit{temporal dynamics} of environments from a set of videos, $\bigcup\limits_{i=1}^{N} \small \bigcup\limits_{t=1}^{T}\{o_{i,t}\}$.
\item \textbf{Demonstration:} These methods learn \textit{task-relevant information} such as identifying agents and enemies from a set of observation-action pairs, $\bigcup\limits_{i=1}^{N}\bigcup\limits_{t=1}^{T}\{o_{i,t}, a_{i,t}\}$.
\item \textbf{Trajectory:} These methods learn \textit{richer task-relevant information} such as
the value of states and actions from a set of observation-action-reward triplets, $\bigcup\limits_{i=1}^{N}\bigcup\limits_{t=1}^{T}\{o_{i,t}, a_{i,t}, r_{i,t}\}$.
\end{itemize}




\section{Algorithms}
\label{section:algorithms}

This section outlines the pre-training algorithms we study, divided into four categories described in Section \ref{subsection:method_class}. Some algorithms were simplified to fit into our framework; we annotate such algorithms with a dagger($\dagger$) to avoid confusion. For further details, please refer to Appendix \ref{appendix:baselines}.

\subsection{No pre-training}

\textbf{Random:} This approach involves fine-tuning a model that has been randomly initialized, with its encoder kept frozen throughout the fine-tuning process. 

\textbf{E2E:} In contrast to the Random method, the End-to-End approach involves fine-tuning a randomly initialized model without any frozen components. In addition to ResNet-50, we also evaluate a 3-layer-CNN based model \cite{mnih2015dqn}, which is known to excel in many RL environments.



\subsection{Learning from Image}
\label{sub_section:learning_from_image}

\textbf{CURL:} Contrastive Unsupervised Reinforcement Learning \cite{laskin2020curl} learns the spatial feature of images using augmentation functions and InfoNCE loss. It operates by ensuring that two augmented instances of the same image are encoded similarly in latent space.


\textbf{MAE:} Masked Autoencoder \citep{he2022mae} learns the spatial structure of images by reconstructing masked images with transformer encoder-decoder architecture. Since we employ a convolutional architecture in this study, we adapt this approach by masking pixels in the convolutional feature map, inspired by \citet{seo2023mwm,xiao2021early}.

\subsection{Learning from Video}
\label{sub_section:learning_from_video}

\textbf{ATC:} Augmented Temporal Contrast \citep{stooke2021atc} learns the temporal dynamics of videos using InfoNCE loss. 
This involves closely aligning an image with its future image in a latent space while maintaining distinctiveness from the other unrelated images.

\textbf{SiamMAE:} Siamese Masked Autoencoder \cite{gupta2023siammae} is an extension of MAE to videos. This variant uses a Siamese architecture and an asymmetric masking strategy for temporal information extraction. In line with MAE, we apply masking to convolutional features.



\textbf{R3M$^{\dagger}$:} Reusable Representations for Robot Manipulation \cite{nair2022r3m} uses multiple losses to learn from human demonstration videos with diverse tasks. 
In our study, we focus on the time contrastive loss of R3M. While akin to ATC, R3M gives an additional task of differentiating between 'near-future' and 'far-future' images.

\subsection{Learning from Demonstration}
\label{sub_section:learning_from_demonstration}

\textbf{BC:} Behavioral Cloning \cite{pomerleau1991bc} learns to imitate the demonstrations by predicting actions from observations. This method acquires task-specific information from the dataset, as noted by \citet{islam2022acro}.

\textbf{SPR:} Self-Predictive Representations \citep{schwarzer2020spr} learn the dynamics of environments by recursively predicting future observations in a latent space.
We employ a recurrent neural network for future prediction and a momentum network for encoding target observations. 

\textbf{IDM:} Inverse Dynamics Modeling \cite{christiano2016idm} learns to predict the action between consecutive observations.
Similar to BC, IDM focuses on learning task-relevant information as it seeks to understand the cause-and-effect dynamics within the environments \citep{islam2022acro}.

\textbf{SPR+IDM:} Inspired by \citet{schwarzer2021sgi}, we investigate whether the combination of SPR and IDM can further improve the agent's performance.




\subsection{Learning from Trajectory}
\label{sub_section:learning_from_trajectory}

\textbf{CQL:} Conservative Q-Learning \cite{kumar2020cql} integrates temporal difference loss with conservative Q-learning loss. Its main objective is to accurately approximate the Bellman target while minimizing the overestimation of actions that are absent in the offline data.
We consider two variants: CQL-M, using mean squared loss, and CQL-D, applying cross-entropy-based distributional backups \cite{bellemare2017c51} for minimizing temporal differences.


\textbf{DT:}
Decision Transformer \cite{chen2021dectransformer} redefines RL as a sequence modeling problem. The network is trained to predict actions based on given states and desired cumulative rewards. During inference, the agent predicts the optimal actions needed to achieve a specified cumulative reward.



\section{Atari Pre-training Benchmark}
\label{section:experimental_setup}
We introduce the Atari Pre-training Benchmark (Atari-PB), a benchmark designed to assess the generalization ability of pre-training methods in vision-based RL.

\subsection{Dataset}
\label{sub_section:dataset}

The dataset for Atari-PB is derived from the DQN-Replay-Dataset \cite{agarwal2020optimistic}. This dataset encompasses training logs from a DQN agent's experiences across 60 Atari games \cite{bellemare2013arcade, machado2018atari}, documented over five distinct runs. Each run is divided into 50 evenly spaced checkpoints, ranging from the initial state (checkpoint 1) to the final state (checkpoint 50). This segmentation allows for precise control over the agent's performance level, thereby influencing the quality of the pre-training data.

To construct a pre-training dataset that reflects real-world complexities as a mixture of suboptimal and optimal behaviors, we selected the first 10 checkpoints from two separate runs. From each checkpoint, we sampled the first 10,000 transitions. Consequently, this process generated 200,000 transitions per game, culminating in a comprehensive dataset of 10 million transitions across 50 games. For more details, please refer to Appendix \ref{appendix:impl_pretrain}.

\subsection{Model}
\label{sub_section:model}

The network architecture of our model is composed of three main components: a backbone for encoding images into features, a neck for converting the features to a low-dimensional latent vector, and a head for mapping the latent vector into policy outputs.


\textbf{Backbone, $f(\cdot)$:}
Our backbone utilizes a modified version of the ResNet-50 architecture \cite{he2016resnet}. It is designed to process input images $\mathbf{o} \in \mathbb{R}^{C \times H \times W}$ and encode them into spatial feature maps $\mathbf{z}=f(\mathbf{o})$, where $\mathbf{z} \in \mathbb{R}^{D_z \times H_z \times W_z}$. Here, $D_z$ represents the output dimension, while $H_z$ and $W_z$ are the dimensions of the feature map's height and width, respectively. 
We replaced the batch normalization \cite{ioffe2015batch} with group normalization \cite{wu2018group}, aiming to address discrepancies in data distributions between pre-training and fine-tuning phases, as discussed in \citet{kumar2022sql, kumar2022ptr}.



\textbf{Neck, $g(\cdot)$:} 
The neck module incorporates a game-specific spatial pooling strategy to manage the variability of in-game elements \cite{kumar2022sql, kumar2022ptr}. It is followed by a 2-layer Multi-Layer Perceptron (MLP) which transforms the spatial features into a low-dimensional latent vector $\mathbf{q}=g(\mathbf{z})$, with $\mathbf{q} \in \mathbb{R}^{D_q}$ and $D_q$ denoting the latent dimension.


\textbf{Head, $h(\cdot)$:} 
The head module employs a game-specific linear layer, allowing diverse policy outputs across different games. This results in the final output $\mathbf{y}=h(\mathbf{q})$, where $\mathbf{y} \in \mathbb{R}^{|A|}$ and $A$ denotes the action space.


\begin{figure}[t]
\begin{center}
\includegraphics[width=0.96\linewidth]{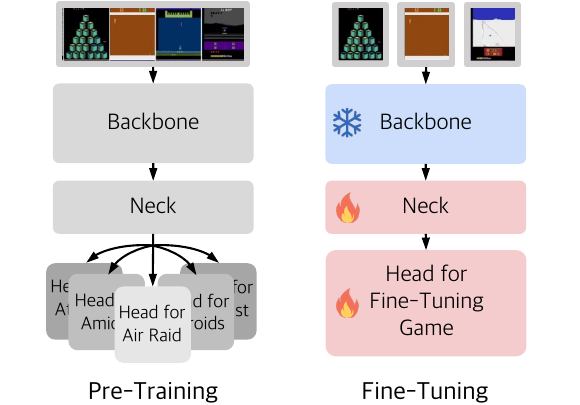}
\end{center}
\caption{\textbf{Experimental Setup.} The model is pre-trained with 50 Atari games in a multi-headed fashion (left), then fine-tuned for each game individually (right).  The snowflake symbol indicates freezing the weights, whereas the fire symbol represents re-initializing and fine-tuning the component.}
\label{figure:introduction}
\end{figure}

\begin{figure*}
\begin{center}
\includegraphics[width=0.99\linewidth]{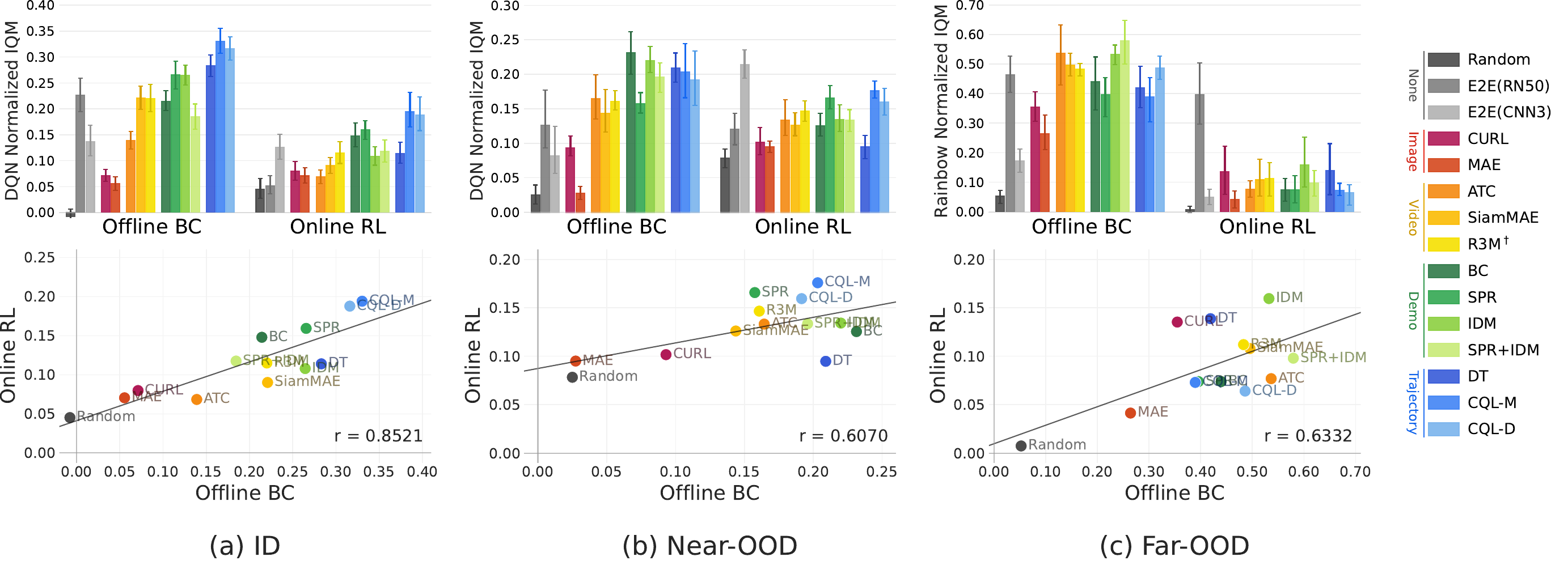}
\end{center}
\caption{\textbf{Main Results.} Performance of each pre-training method after fine-tuning, in different distributions (ID, Near-OOD, Far-OOD) and adaptation scenarios (Offline BC, Online RL).
We report the Inter Quantile Mean (IQM) of normalized scores across three seeds, along with a 95\% confidence interval.
The bars are grouped and color-coded by their categories described in Section \ref{subsection:method_class} for ease of view.
}
\label{figure:main}
\end{figure*}


\subsection{Pre-Training}
\label{sub_section:pre-training}
Following standard training protocols in Atari games, we pre-processed each image by down-sampling to $84 \times 84$ with grey-scaling, then stacked 4 consecutive frames \cite{mnih2015dqn, hessel2018rainbow}. We applied two image augmentations: a random shift followed by intensity jittering \cite{schwarzer2020data, schwarzer2021pretraining}. 
Each model was pre-trained for 100 epochs using an AdamW optimizer \cite{loshchilov2017decoupled} with a batch size of $512$.
We experimented with various learning rates, selecting from the range of $\{1e-3, 3e-4, ..., 3e-5, 1e-6\}$, and adjusted the weight decay within the range of $\{1e-4, 1e-5, 1e-6\}$.


\subsection{Fine-tuning}
\label{sub_section:fine-tuning}

For fine-tuning, we assessed the effectiveness of our pre-trained models in three different categories of environments:

\begin{itemize}[itemsep=0pt, topsep=0pt]
\item \textbf{In-Distribution (ID)}: This category includes the same 50 games used during the pre-training phase. This assessment aims to evaluate model performance in familiar settings. These games include the main task genres of Atari games, such as maze-based, tracking, vertical shooting, and horizontal shooting games. 



\item \textbf{Near-Out-of-Distribution (Near-OOD)}: This group consists of 10 games not used in pre-training but belonging to the same task genres as the ID games. While these games present tasks similar to those in ID, they assess the model's generalization ability on new visual elements and reward structures.


\item \textbf{Far-Out-of-Distribution (Far-OOD)}:  This group extends beyond Near-OOD, featuring 5 games with entirely novel task mechanics. For instance, HumanCannonball introduces projectile motion against gravity, while Klax focuses on color matching and stacking. These games serve as the baselines to understand the pre-trained models' generalization ability in entirely unfamiliar environments and tasks.
\end{itemize}


For these environments, we consider two common adaptation scenarios:

\begin{itemize}[itemsep=1pt, topsep=0pt]
\item \textbf{Offline BC}: 
In this scenario, the model undergoes fine-tuning through behavior cloning using 50,000 frames from expert demonstrations. For ID and Near-OOD games, expert demonstrations were sourced from the final checkpoint of the DQN-Replay-Dataset. For Far-OOD games, we used demonstrations from a Rainbow agent \cite{hessel2018rainbow} trained for 2 million steps. During this process, as depicted in Figure \ref{figure:introduction}, we kept the backbone parameters frozen, while the neck and head components were re-initialized and then fine-tuned for 100 epochs.


\item \textbf{Online RL}: In this scenario, the model is fine-tuned using the Rainbow algorithm \cite{hessel2018rainbow}, with 50,000 interactions in each respective environments. Identical to the Offline BC approach, the backbone of the pre-trained model remains unaltered, while the neck and head are re-initialized and subsequently fine-tuned.


\end{itemize}

To ensure a reliable evaluation, we report the normalized game scores using the Inter Quantile Mean (IQM) methodology \cite{agarwal2021iqm}. This method involves stratified bootstrap sampling and is executed with three different random seeds. For normalizing the scores in ID and Near-OOD environments, we use the DQN scores as documented by \citet{castro18dopamine}. In the case of Far-OOD scenarios, the normalization is based on the final scores of our Rainbow agent, which was trained for 2 million steps to provide the expert dataset for the Offline BC scenario.

\section{Experimental Results}
\label{sec:main_result}

\begin{figure}
\begin{center}
\includegraphics[width=1.0\linewidth]{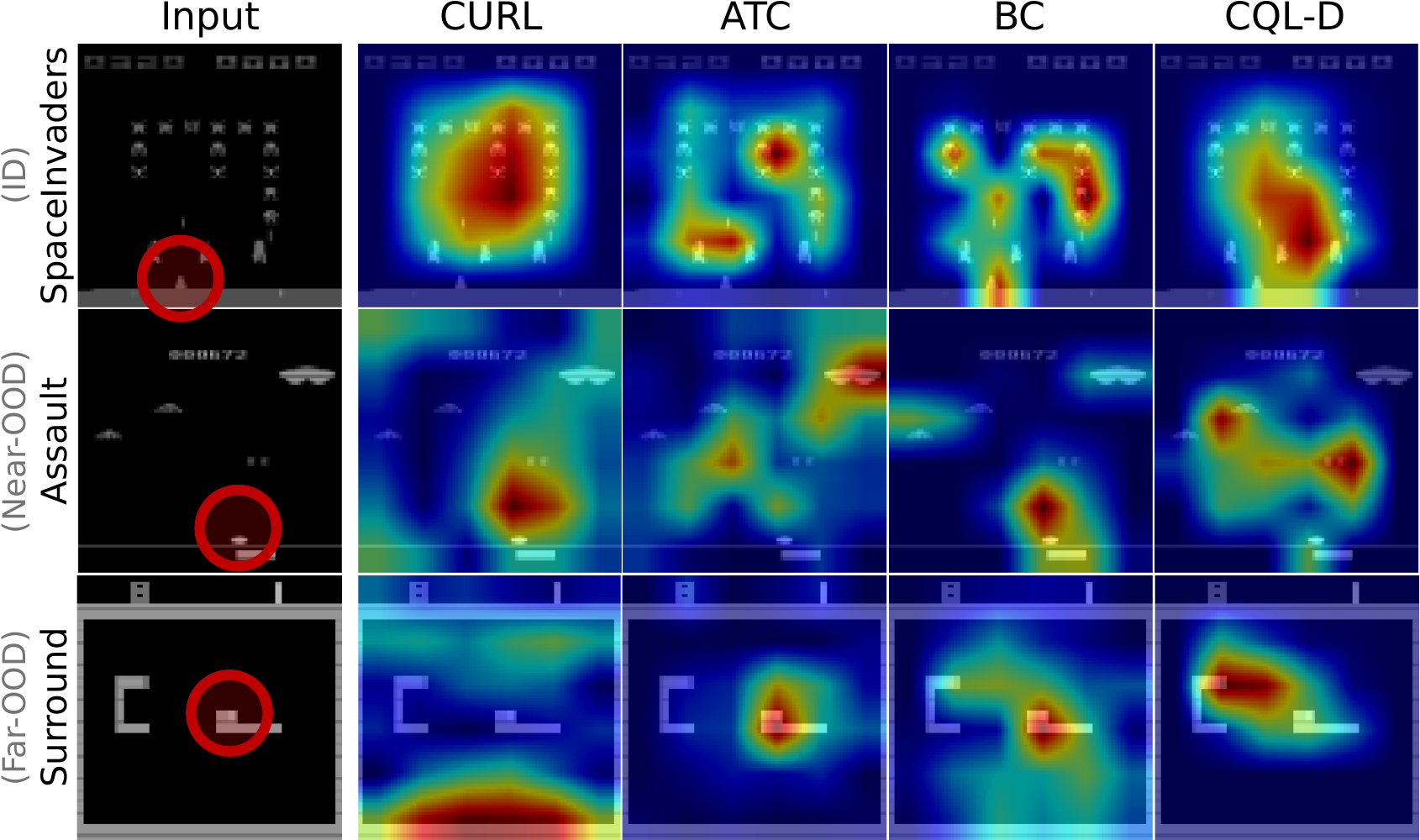}
\end{center}
\caption{\textbf{Qualitative analysis of methods.} EigenCAM visualization of the pre-trained backbones in 3 games: SpaceInvaders (ID), Assault (Near-OOD), and Surround (Far-OOD).
The agents are marked in red circles for each game (first column).
We chose one representative method for each algorithm class.}
\label{figure:main_eigencam}
\end{figure}

In this section, we present our main experimental results, as illustrated in Figure \ref{figure:main} and \ref{figure:main_eigencam}.
Instead of merely providing a ranking of different pre-training methods, we focus on discerning specific trends and patterns that manifest across different downstream distributions and adaptation scenarios.

In summary, a distinctive pattern emerged between algorithms that learn task-agnostic information (using images or videos), and those that learn task-relevent information (using demonstrations or trajectories). While task-agnostic knowledge consistently improved performance across all distributions, task-specific knowledge showed limited robustness against task distribution shifts. We further elaborate on our findings through several key observations.

\textit{O1: Learning task-agnostic information from images and videos significantly enhance performance across ID, Near-OOD, and Far-OOD environments.}

Our findings, as depicted in Figure \ref{figure:main}, reveal that image-based pre-training methods like CURL and MAE consistently surpassed the Random baseline (i.e., a frozen backbone with randomly initialized weights). This highlights the importance of extracting spatial information (e.g., object size, shape, location) for an effective adaptation to ID environments and generalization to OOD environments.

Moreover, video-based pre-training methods (ATC, SiamMAE, R3M$^\dagger$) showed enhanced performance over image-based approaches, underscoring the significance of capturing both spatial and temporal information (e.g., objects' moving orientation and speed). Notably, these video-pretrained models slightly surpassed end-to-end (E2E) fine-tuning from scratch, despite having their backbone parameters frozen during fine-tuning. We believe that allowing end-to-end fine-tuning of video-based models will further enhance their performance, widening the performance gap compared to the E2E approach.

These findings support the emerging trend of incorporating images and videos in pre-training to achieve better generalization across various control tasks \citep{majumdar2023vc1, bhateja2023vptr}.

\textit{O2: Learning task-relevant information from demonstrations further enhances ID and Near-OOD performance, but provides marginal improvements in Far-OOD performance.}

Incorporating actions into pre-training objectives (BC, SPR, IDM, and SPR+IDM) led to mixed performance improvements. While task-specific knowledge benefited performance in ID and Near-OOD settings, a decline was noted in Far-OOD environments. 
This variation likely arose from the task similarities between ID and Near-OOD environments, in contrast to the distinct differences in tasks between ID and Far-OOD. Given that ID and Near-OOD games typically fall into the four main genres, it's plausible that the task knowledge from ID environments can be applied to Near-OOD environments, but not Far-OOD.

For example, as illustrated in the first column of Figure \ref{figure:main_eigencam}, the model pre-trained on SpaceInvaders (ID) learns task-specific knowledge such as agent and enemy locations as well as bullet movement direction, which are characteristic of the vertical shooting genre.
This knowledge helps the model to quickly identify similar gameplay elements in Assault (Near-OOD), despite not having been exposed to the game during pre-training. Conversely, task-specific knowledge is rendered ineffective in Surround (Far-OOD), with its unique task of avoiding trails and distinct agent locations. This difference shows the difficulty in applying pre-learned task knowledge to environments with considerably different mechanisms.

\textit{O3: Learning reward-specific information from trajectories yields the best ID performance, while it shows limited generalization gains in Near-OOD and Far-OOD environments.}

Integrating reward-specific information from trajectories (DT, CQL-M, CQL-D) led to superior success in ID environments. Nonetheless, their effectiveness diminished in Near-OOD and Far-OOD environments, where they lagged behind demonstration-based and video-based approaches. 
This shows that extracting reward-specific knowledge (e.g., rewarded for shooting enemies) enhances performance in familiar settings, but fails to generalize across environments with different reward functions (e.g., rewarded for shooting enemies but penalized for missing) or tasks (e.g., receiving rewards when surrounding the enemy), indicating their limitations for broader generalization.


To further validate our observations, we conducted a qualitative analysis of the pre-trained models with Eigen-CAM \cite{muhammad2020eigencam}, which are illustrated in Figure \ref{figure:main_eigencam}.
In ID games like SpaceInvaders, we found that unlike video-based methods (ATC), the demonstration or trajectory-based methods (BC, CQL-D) concentrated on capturing the agent, a critical object of the task.
Interestingly, an identical pattern was observed in Near-OOD environments like Assault, fortifying our conjecture that the task-specific knowledge acquired during pre-training can be transferred to visually distinct environments. For an extended discussion and analysis, refer to Appendix \ref{supp:eigen_cam}.

\textit{O4: Effective adaptation in one scenario correlates to effective adaptation in the other.}

Figure \ref{figure:main} shows a strong correlation between performances in offline behavioral cloning and online reinforcement learning, with Pearson correlation coefficients of 0.85 in ID, 0.61 in Near-OOD, and 0.63 in Far-OOD environments. 

This suggests that a well-pre-trained model can yield versatile representations applicable to a wide range of scenarios, not limited to specific adaptation algorithms \cite{parisi2022pvr, majumdar2023cortexbench}.

\begin{figure}[t]
\begin{center}
\includegraphics[width=0.99\linewidth]{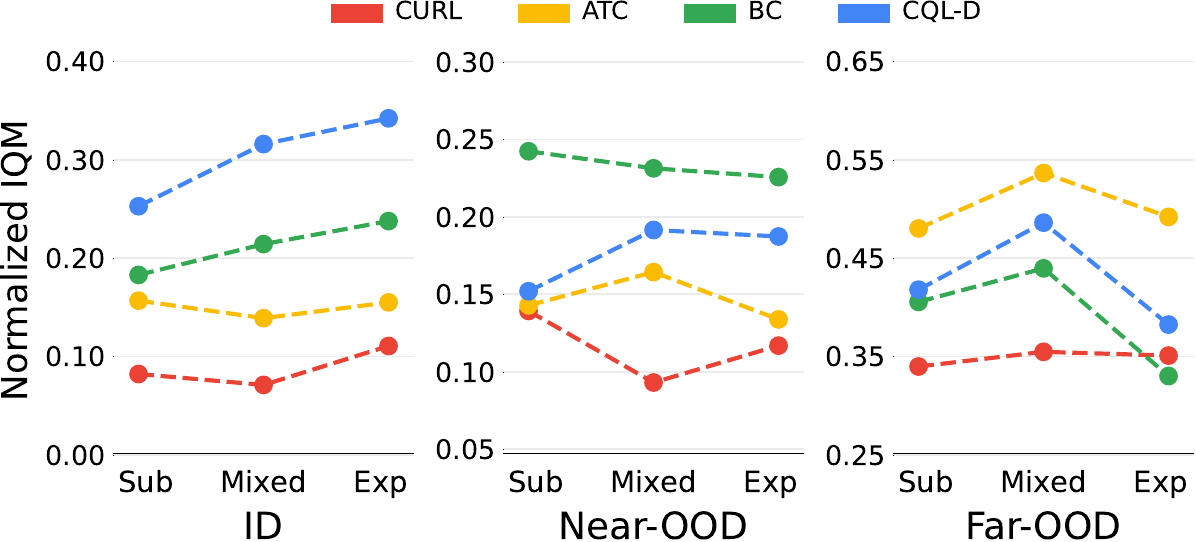}
\end{center}
\vspace{-2mm}
\caption{\textbf{Effect of data optimality.} Offline BC performance of algorithms after pre-training with datasets of differing optimality. We control the dataset optimality via the proficiency of the policy that created it, which in turn can be controlled by choosing different checkpoints of the DQN-Replay-Dataset \cite{agarwal2020optimistic}.}
\label{figure:data_suboptimality}
\end{figure}

\textit{O5: Miscellaneous Remarks}

Inspired by \citet{taiga2022investigating}, we have conducted an additional experiment to fine-tune our pre-trained models on ID environments with different game modes \cite{machado2018atari}. Detailed protocol and results can be found in Appendix \ref{supp:gamemode}. 
As shown in Figure \ref{fig:change_game_result}, the results were similar to that of ID in our main experiments. We believe that simply changing the game mode did not sufficiently alter the reward functions and thus led to similar tendencies.

    Additionally, we noticed that in end-to-end fine-tuning for Online RL, the smaller 3-layer CNN agent (CNN3) often outperforms the larger ResNet-50 agent (RN50). This finding is consistent with prior research, which suggests that deep and large models are prone to overfitting and are more vulnerable to the non-stationary nature of Online RL \citep{lee2024hare}. Many studies have proposed techniques to mitigate this issue \citep{nikishin2022primacy, lee2024plastic, farebrother2024stop}, suggesting that applying such methods could make larger architectures more competitive.

Finally, we have found an interesting relationship between object size and the performance of mask-based algorithms (MAE, SiamMAE). Compared to other pre-training methods like CURL and ATC, these methods excel in environments with large objects but underperform in games with tiny objects. For more details, please refer to Appendix \ref{supp:obj_size}.

\section{Ablation Studies}

In our ablation studies, we assess the effects of variations in data optimality, size, and model size on pre-training methods. We selected CURL (image), ATC (video), BC (demonstration), and CQL-D (trajectory) as representative methods for comparison. Unless otherwise noted, the experimental setup remains identical to our main experiments.

\subsection{Data Optimality}
\label{subsection:suboptimality}

A key factor in building pre-training datasets for RL is data optimality. We investigated the impact of data optimality using three distinct datasets from the DQN-Replay-Dataset:


\begin{itemize}[itemsep=0pt, topsep=0pt]
\item \textbf{Suboptimal}:  Data from the first and second checkpoint of each game, each with 50,000 interactions.
\item \textbf{Mixed}: Data from the first ten checkpoints per game, each with 10,000 interactions (same as our main setup).
\item \textbf{Expert}:  Data from the ninth and tenth checkpoint of each game, each with 50,000 interactions.
\end{itemize}

\begin{figure}[t]
\begin{center}
\includegraphics[width=0.99\linewidth]{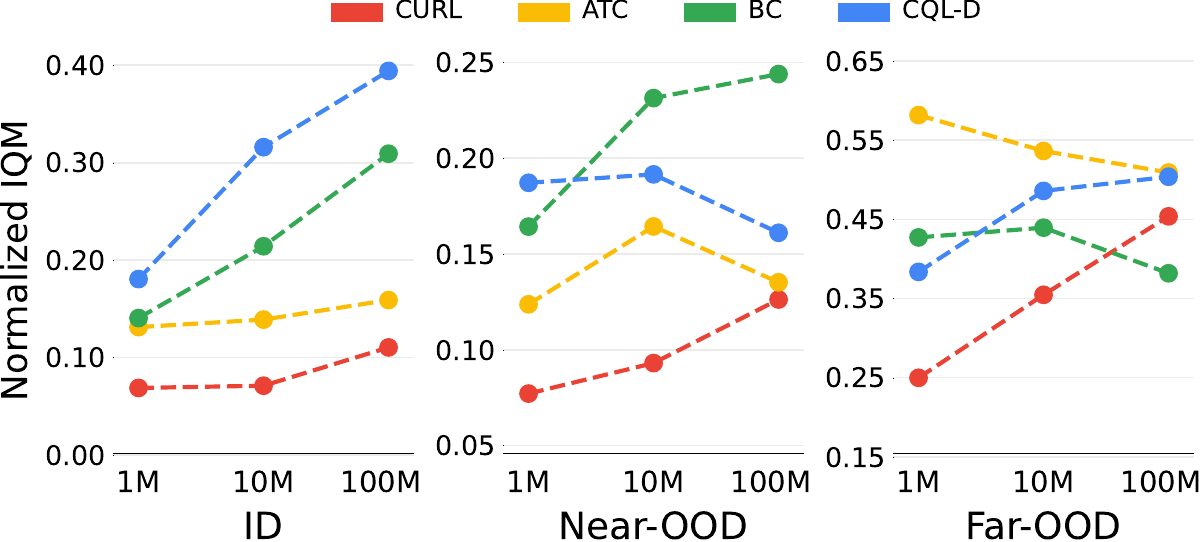}
\end{center}
\caption{\textbf{Effect of Dataset Size.} We measure the Offline BC performance after pre-training with datasets of varying sizes. All datasets were derived from the same DQN-Replay-Dataset runs and checkpoints, with different numbers of transitions sampled based on dataset size.
}
\label{figure:data_scale}
\end{figure}

\textit{O6: Using optimal data enhances ID adaptation but falters generalization in Near-OOD and Far-OOD environments.}

Figure \ref{figure:data_suboptimality} shows that using optimal data does not guarantee improved performance in downstream environments. While moving from a Mixed dataset to an Expert dataset enhances performance in ID environments, its effectiveness was limited in Near-OOD and Far-OOD environments.

Optimal gameplay in Atari games usually follows repetitive patterns, resulting in limited diversity in Expert dataset. Such uniformity likely hindered the pre-trained model's ability to generalize to unfamiliar objects, by making it overly tailored to the objects encountered during pre-training. On the other hand, models pre-trained on datasets that blend optimal and suboptimal transitions (Mixed) showed enhanced generalization capabilities, particularly in Far-OOD environments. This improvement underscores the significance of dataset diversity for achieving effective generalization, a principle supported by recent research \cite{taiga2022investigating}.

\begin{figure}[t]
\begin{center}
\includegraphics[width=0.99\linewidth]{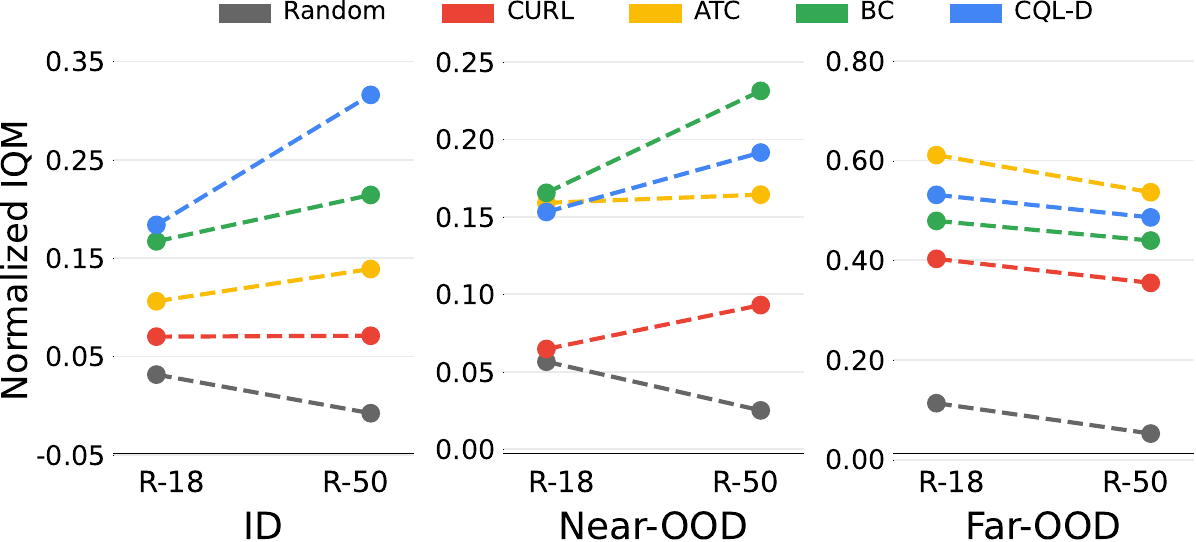}
\end{center}
\caption{\textbf{Effect of model size.} Offline BC performance of algorithms utilizing different-sized models.
We scaled the depth of models by the number of layers, and the width by the number of channels. Here, we also provide the Random performance of the modified models as a baseline.}
\label{figure:model_scale}
\end{figure}

\subsection{Data Size}
\label{subsection:data_scale}

The scalability of pre-training methods can be significantly influenced by the size of the pre-training dataset. We explored this by using datasets of varying sizes: 1M, 10M, and 100M transitions, each derived from the same runs and checkpoints but adjusted to their respective sizes.

\begin{itemize}[itemsep=0pt, topsep=0pt]
    \item \textbf{1M}: Consists of 1,000 initial transitions, sampled from the 10 initial checkpoints of each run. 
    \item \textbf{10M}: Our standard setup, with 10,000 initial transitions from each checkpoint.
    \item \textbf{100M}: The largest dataset, containing 100,000 initial transitions from each checkpoint.
\end{itemize}

Models were pre-trained over 100 epochs for 1M and 10M datasets. However, due to the substantial size of the 100M dataset, pre-training was limited to 10 epochs.

\textit{O7: Larger dataset enhances ID adaptation but shows mixed effects on Near-OOD and Far-OOD generalization.} 

Figure \ref{figure:data_scale} shows that larger datasets improved performance in ID settings, especially for methods involving demonstrations or trajectories. However, the benefits were less predictable in Near-OOD and Far-OOD environments, possibly because the model is relatively too small to fully encode all the information within large datasets in 10 epochs.

Consistent with our earlier findings in \textit{O1}, \textit{O2}, and \textit{O3}, demonstration-based methods excelled in Near-OOD settings, while video-based methods were more effective in Far-OOD scenarios. This reiterates our findings that learning task-agnostic features improves generalization across distributions, but the benefits of learning task-relevant features diminish as task shifts increase.

\subsection{Model Size}
\label{subsection:model_scale}

In the subsequent analysis, we explore how pre-training methods scale with changes in model size:

\begin{itemize}[itemsep=0pt, topsep=0pt]
    \item \textbf{R-18}: A scaled-down model with reduced depth (18 layers) and halved channel widths, compared to R-50.
    \item \textbf{R-50}: A standard model architecture used in our study.
\end{itemize}

\textit{O8: Larger model provides consistent benefits in ID and Near-OOD, while its improvement is unclear in Far-OOD. }

When comparing ResNet-18 to ResNet-50, we found that larger models improve results in ID and Near-OOD scenarios. While broader computer vision research suggests larger models usually offer better generalization to out-of-distribution (OOD) environments, this trend was not clearly observed in our Far-OOD experiments.

Nevertheless, the experimental results aligned with our prior observations (\textit{O1}, \textit{O2}, and \textit{O3}), indicating that demonstration-based methods perform better in Near-OOD environments, and video-based methods excel in Far-OOD environments, irrespective to their model size.

\section{Conclusion and Future Work}

In this study, we examined how different pre-training objectives affect agents' generalization capabilities in vision-based reinforcement learning (RL). Our findings indicate that learning task-agnostic features, such as spatial features (e.g., object locations and shapes) and temporal features (e.g., speed and direction of moving objects), enhances generalization across various visual and task distribution shifts. In contrast, learning task-specific features, through actions (e.g., locations of agents and related objects) or reward structures (e.g., identifying beneficial or detrimental actions), improves performance in similar tasks but offers limited benefits in divergent task distributions.

Our results align with prior theoretical works demonstrating the benefits of learning temporal structures in environments for acquiring optimal feature representations \cite{bellemare2019geometric, lan2022generalization}. In this work, we provide empirical evidences supporting the notion that learning temporal structures within environments can improve generalization across various types of task shifts.

The RL community has recently introduced diverse datasets covering a wide range of human and robotic behaviors \cite{padalkar2023rtx, grauman2022ego4d}. Our results suggest that pre-training with both task-agnostic and task-specific knowledge offers distinct benefits: one enhances generalization to different shifts, while the other excels in similar environments. Therefore, a promising future direction is to develop architectures or learning objectives that can decouple task-agnostic and task-specific features, allowing their use based on specific purposes \citep{wang2022vrl3, bhateja2023vptr}. We hope our investigation provides valuable insights for advancing vision-based RL.

\section*{Acknowledgements}

This work was supported by Institute for Information \& communications Technology Promotion(IITP) grant funded by the Korea government(MSIT) (No.RS-2019-II190075 Artificial Intelligence Graduate School Program(KAIST)), the National Research Foundation of Korea (NRF) grant funded by the Korea government (MSIT) (No. NRF-2022R1A2B5B02001913).

\section*{Impact Statement}

This paper investigates the relationship between pre-training objectives and their generalization capabilities across different downstream environments. Our findings provide valuable insights into how various learning objectives influence model generalization, aiding the development of large, foundational vision-based agents.

We also acknowledge the potential risks associated with rapid advancements in these foundational models, especially in robotic control. Discussions for preventing misuse and maintaining high ethical standards are essential to ensure these technologies benefit society and do not cause harm.

\bibliography{main}

\begin{thebibliography}{101}
\providecommand{\natexlab}[1]{#1}
\providecommand{\url}[1]{\texttt{#1}}
\expandafter\ifx\csname urlstyle\endcsname\relax
  \providecommand{\doi}[1]{doi: #1}\else
  \providecommand{\doi}{doi: \begingroup \urlstyle{rm}\Url}\fi

\bibitem[Agarwal et~al.(2020)Agarwal, Schuurmans, and Norouzi]{agarwal2020optimistic}
Agarwal, R., Schuurmans, D., and Norouzi, M.
\newblock An optimistic perspective on offline reinforcement learning.
\newblock \emph{Proc. the International Conference on Machine Learning (ICML)}, 2020.

\bibitem[Agarwal et~al.(2021)Agarwal, Schwarzer, Castro, Courville, and Bellemare]{agarwal2021iqm}
Agarwal, R., Schwarzer, M., Castro, P.~S., Courville, A.~C., and Bellemare, M.
\newblock Deep reinforcement learning at the edge of the statistical precipice.
\newblock \emph{Proc. the Advances in Neural Information Processing Systems (NeurIPS)}, 34:\penalty0 29304--29320, 2021.

\bibitem[Arora et~al.(2020)Arora, Du, Kakade, Luo, and Saunshi]{arora2020provablebc}
Arora, S., Du, S., Kakade, S., Luo, Y., and Saunshi, N.
\newblock Provable representation learning for imitation learning via bi-level optimization.
\newblock In \emph{International Conference on Machine Learning}, 2020.

\bibitem[Baker et~al.(2022)Baker, Akkaya, Zhokhov, Huizinga, Tang, Ecoffet, Houghton, Sampedro, and Clune]{baker2022vpt}
Baker, B., Akkaya, I., Zhokhov, P., Huizinga, J., Tang, J., Ecoffet, A., Houghton, B., Sampedro, R., and Clune, J.
\newblock Video pretraining (vpt): Learning to act by watching unlabeled online videos.
\newblock \emph{Proc. the Advances in Neural Information Processing Systems (NeurIPS)}, 2022.

\bibitem[Bellemare et~al.(2019)Bellemare, Dabney, Dadashi, Ali~Taiga, Castro, Le~Roux, Schuurmans, Lattimore, and Lyle]{bellemare2019geometric}
Bellemare, M., Dabney, W., Dadashi, R., Ali~Taiga, A., Castro, P.~S., Le~Roux, N., Schuurmans, D., Lattimore, T., and Lyle, C.
\newblock A geometric perspective on optimal representations for reinforcement learning.
\newblock \emph{Advances in neural information processing systems}, 32, 2019.

\bibitem[Bellemare et~al.(2013)Bellemare, Naddaf, Veness, and Bowling]{bellemare2013arcade}
Bellemare, M.~G., Naddaf, Y., Veness, J., and Bowling, M.
\newblock The arcade learning environment: An evaluation platform for general agents.
\newblock \emph{Journal of Artificial Intelligence Research}, 2013.

\bibitem[Bellemare et~al.(2017{\natexlab{a}})Bellemare, Dabney, and Munos]{bellemare2017c51}
Bellemare, M.~G., Dabney, W., and Munos, R.
\newblock A distributional perspective on reinforcement learning.
\newblock In \emph{Proc. the International Conference on Machine Learning (ICML)}, pp.\  449--458. PMLR, 2017{\natexlab{a}}.

\bibitem[Bellemare et~al.(2017{\natexlab{b}})Bellemare, Dabney, and Munos]{bellemare2017distributional}
Bellemare, M.~G., Dabney, W., and Munos, R.
\newblock A distributional perspective on reinforcement learning.
\newblock 2017{\natexlab{b}}.

\bibitem[Bhateja et~al.(2023)Bhateja, Guo, Ghosh, Singh, Tomar, Vuong, Chebotar, Levine, and Kumar]{bhateja2023vptr}
Bhateja, C., Guo, D., Ghosh, D., Singh, A., Tomar, M., Vuong, Q., Chebotar, Y., Levine, S., and Kumar, A.
\newblock Robotic offline rl from internet videos via value-function pre-training.
\newblock \emph{Proc. the Advances in Neural Information Processing Systems (NeurIPS)}, 2023.

\bibitem[Brandfonbrener et~al.(2023)Brandfonbrener, Nachum, and Bruna]{brandfonbrener2023idm-theory}
Brandfonbrener, D., Nachum, O., and Bruna, J.
\newblock Inverse dynamics pretraining learns good representations for multitask imitation.
\newblock \emph{Proc. the Advances in Neural Information Processing Systems (NeurIPS)}, 2023.

\bibitem[Bubeck et~al.(2023)Bubeck, Chandrasekaran, Eldan, Gehrke, Horvitz, Kamar, Lee, Lee, Li, Lundberg, et~al.]{bubeck2023gpt4}
Bubeck, S., Chandrasekaran, V., Eldan, R., Gehrke, J., Horvitz, E., Kamar, E., Lee, P., Lee, Y.~T., Li, Y., Lundberg, S., et~al.
\newblock Sparks of artificial general intelligence: Early experiments with gpt-4.
\newblock \emph{arXiv preprint arXiv:2303.12712}, 2023.

\bibitem[Caluwaerts et~al.(2023)Caluwaerts, Iscen, Kew, Yu, Zhang, Freeman, Lee, Lee, Saliceti, Zhuang, et~al.]{caluwaerts2023barkour}
Caluwaerts, K., Iscen, A., Kew, J.~C., Yu, W., Zhang, T., Freeman, D., Lee, K.-H., Lee, L., Saliceti, S., Zhuang, V., et~al.
\newblock Barkour: Benchmarking animal-level agility with quadruped robots.
\newblock \emph{arXiv preprint arXiv:2305.14654}, 2023.

\bibitem[Castro et~al.(2018)Castro, Moitra, Gelada, Kumar, and Bellemare]{castro18dopamine}
Castro, P.~S., Moitra, S., Gelada, C., Kumar, S., and Bellemare, M.~G.
\newblock Dopamine: {A} {R}esearch {F}ramework for {D}eep {R}einforcement {L}earning.
\newblock 2018.
\newblock URL \url{http://arxiv.org/abs/1812.06110}.

\bibitem[Chen et~al.(2021)Chen, Lu, Rajeswaran, Lee, Grover, Laskin, Abbeel, Srinivas, and Mordatch]{chen2021dectransformer}
Chen, L., Lu, K., Rajeswaran, A., Lee, K., Grover, A., Laskin, M., Abbeel, P., Srinivas, A., and Mordatch, I.
\newblock Decision transformer: Reinforcement learning via sequence modeling.
\newblock \emph{Proc. the Advances in Neural Information Processing Systems (NeurIPS)}, 2021.

\bibitem[Chen et~al.(2020{\natexlab{a}})Chen, Kornblith, Norouzi, and Hinton]{chen2020simclr}
Chen, T., Kornblith, S., Norouzi, M., and Hinton, G.
\newblock A simple framework for contrastive learning of visual representations.
\newblock In \emph{Proc. the International Conference on Machine Learning (ICML)}, pp.\  1597--1607. PMLR, 2020{\natexlab{a}}.

\bibitem[Chen \& He(2021)Chen and He]{chen2021simsiam}
Chen, X. and He, K.
\newblock Exploring simple siamese representation learning.
\newblock In \emph{Proc. of the IEEE Conference on Computer Vision and Pattern Recognition (CVPR)}, pp.\  15750--15758, 2021.

\bibitem[Chen et~al.(2020{\natexlab{b}})Chen, Fan, Girshick, and He]{chen2020mocov2}
Chen, X., Fan, H., Girshick, R., and He, K.
\newblock Improved baselines with momentum contrastive learning.
\newblock \emph{arXiv preprint arXiv:2003.04297}, 2020{\natexlab{b}}.

\bibitem[Christiano et~al.(2016)Christiano, Shah, Mordatch, Schneider, Blackwell, Tobin, Abbeel, and Zaremba]{christiano2016idm}
Christiano, P., Shah, Z., Mordatch, I., Schneider, J., Blackwell, T., Tobin, J., Abbeel, P., and Zaremba, W.
\newblock Transfer from simulation to real world through learning deep inverse dynamics model.
\newblock \emph{arXiv preprint arXiv:1610.03518}, 2016.

\bibitem[Cobbe et~al.(2020)Cobbe, Hesse, Hilton, and Schulman]{cobbe2020procgen}
Cobbe, K., Hesse, C., Hilton, J., and Schulman, J.
\newblock Leveraging procedural generation to benchmark reinforcement learning.
\newblock In \emph{Proc. the International Conference on Machine Learning (ICML)}, pp.\  2048--2056. PMLR, 2020.

\bibitem[Deng et~al.(2009)Deng, Dong, Socher, Li, Li, and Fei-Fei]{deng2009imagenet}
Deng, J., Dong, W., Socher, R., Li, L.-J., Li, K., and Fei-Fei, L.
\newblock Imagenet: A large-scale hierarchical image database.
\newblock In \emph{2009 IEEE conference on computer vision and pattern recognition}, pp.\  248--255. Ieee, 2009.

\bibitem[Dosovitskiy et~al.(2017)Dosovitskiy, Ros, Codevilla, Lopez, and Koltun]{dosovitskiy2017carla}
Dosovitskiy, A., Ros, G., Codevilla, F., Lopez, A., and Koltun, V.
\newblock Carla: An open urban driving simulator.
\newblock In \emph{Conference on robot learning}, pp.\  1--16. PMLR, 2017.

\bibitem[Dosovitskiy et~al.(2020)Dosovitskiy, Beyer, Kolesnikov, Weissenborn, Zhai, Unterthiner, Dehghani, Minderer, Heigold, Gelly, et~al.]{dosovitskiy2020vit}
Dosovitskiy, A., Beyer, L., Kolesnikov, A., Weissenborn, D., Zhai, X., Unterthiner, T., Dehghani, M., Minderer, M., Heigold, G., Gelly, S., et~al.
\newblock An image is worth 16x16 words: Transformers for image recognition at scale.
\newblock \emph{arXiv preprint arXiv:2010.11929}, 2020.

\bibitem[Farebrother et~al.(2018)Farebrother, Machado, and Bowling]{farebrother2018generalization}
Farebrother, J., Machado, M.~C., and Bowling, M.
\newblock Generalization and regularization in dqn.
\newblock \emph{arXiv preprint arXiv:1810.00123}, 2018.

\bibitem[Farebrother et~al.(2024)Farebrother, Orbay, Vuong, Ta{\"\i}ga, Chebotar, Xiao, Irpan, Levine, Castro, Faust, et~al.]{farebrother2024stop}
Farebrother, J., Orbay, J., Vuong, Q., Ta{\"\i}ga, A.~A., Chebotar, Y., Xiao, T., Irpan, A., Levine, S., Castro, P.~S., Faust, A., et~al.
\newblock Stop regressing: Training value functions via classification for scalable deep rl.
\newblock \emph{arXiv preprint arXiv:2403.03950}, 2024.

\bibitem[Feichtenhofer et~al.(2022)Feichtenhofer, Li, He, et~al.]{feichtenhofer2022mae-st}
Feichtenhofer, C., Li, Y., He, K., et~al.
\newblock Masked autoencoders as spatiotemporal learners.
\newblock \emph{Advances in neural information processing systems}, 35:\penalty0 35946--35958, 2022.

\bibitem[Fujimoto et~al.(2019)Fujimoto, Meger, and Precup]{fujimoto2019bcq}
Fujimoto, S., Meger, D., and Precup, D.
\newblock Off-policy deep reinforcement learning without exploration.
\newblock In \emph{Proc. the International Conference on Machine Learning (ICML)}, 2019.

\bibitem[Grauman et~al.(2022)Grauman, Westbury, Byrne, Chavis, Furnari, Girdhar, Hamburger, Jiang, Liu, Liu, et~al.]{grauman2022ego4d}
Grauman, K., Westbury, A., Byrne, E., Chavis, Z., Furnari, A., Girdhar, R., Hamburger, J., Jiang, H., Liu, M., Liu, X., et~al.
\newblock Ego4d: Around the world in 3,000 hours of egocentric video.
\newblock In \emph{Proceedings of the IEEE/CVF Conference on Computer Vision and Pattern Recognition}, pp.\  18995--19012, 2022.

\bibitem[Grill et~al.(2020)Grill, Strub, Altch{\'e}, Tallec, Richemond, Buchatskaya, Doersch, Avila~Pires, Guo, Gheshlaghi~Azar, et~al.]{grill2020byol}
Grill, J.-B., Strub, F., Altch{\'e}, F., Tallec, C., Richemond, P., Buchatskaya, E., Doersch, C., Avila~Pires, B., Guo, Z., Gheshlaghi~Azar, M., et~al.
\newblock Bootstrap your own latent-a new approach to self-supervised learning.
\newblock \emph{Proc. the Advances in Neural Information Processing Systems (NeurIPS)}, 2020.

\bibitem[Gupta et~al.(2023)Gupta, Wu, Deng, and Fei-Fei]{gupta2023siammae}
Gupta, A., Wu, J., Deng, J., and Fei-Fei, L.
\newblock Siamese masked autoencoders.
\newblock \emph{arXiv preprint arXiv:2305.14344}, 2023.

\bibitem[Hafner et~al.(2019{\natexlab{a}})Hafner, Lillicrap, Ba, and Norouzi]{hafner2019dreamer}
Hafner, D., Lillicrap, T., Ba, J., and Norouzi, M.
\newblock Dream to control: Learning behaviors by latent imagination.
\newblock \emph{arXiv preprint arXiv:1912.01603}, 2019{\natexlab{a}}.

\bibitem[Hafner et~al.(2019{\natexlab{b}})Hafner, Lillicrap, Fischer, Villegas, Ha, Lee, and Davidson]{hafner2019rssm}
Hafner, D., Lillicrap, T., Fischer, I., Villegas, R., Ha, D., Lee, H., and Davidson, J.
\newblock Learning latent dynamics for planning from pixels.
\newblock In \emph{International conference on machine learning}, pp.\  2555--2565. PMLR, 2019{\natexlab{b}}.

\bibitem[Hafner et~al.(2023)Hafner, Pasukonis, Ba, and Lillicrap]{hafner2023dreamerv3}
Hafner, D., Pasukonis, J., Ba, J., and Lillicrap, T.
\newblock Mastering diverse domains through world models.
\newblock \emph{arXiv preprint arXiv:2301.04104}, 2023.

\bibitem[Han et~al.(2019)Han, Xie, and Zisserman]{han2019dense_cpc}
Han, T., Xie, W., and Zisserman, A.
\newblock Video representation learning by dense predictive coding.
\newblock In \emph{Proc. of the IEEE International Conference on Computer Cision (ICCV)}, pp.\  0--0, 2019.

\bibitem[Hansen \& Wang(2021)Hansen and Wang]{hansen2021dmcgb}
Hansen, N. and Wang, X.
\newblock Generalization in reinforcement learning by soft data augmentation.
\newblock In \emph{2021 IEEE International Conference on Robotics and Automation (ICRA)}, pp.\  13611--13617. IEEE, 2021.

\bibitem[He et~al.(2016)He, Zhang, Ren, and Sun]{he2016resnet}
He, K., Zhang, X., Ren, S., and Sun, J.
\newblock Deep residual learning for image recognition.
\newblock In \emph{Proc. of the IEEE Conference on Computer Vision and Pattern Recognition (CVPR)}, 2016.

\bibitem[He et~al.(2020)He, Fan, Wu, Xie, and Girshick]{he2020mocov1}
He, K., Fan, H., Wu, Y., Xie, S., and Girshick, R.
\newblock Momentum contrast for unsupervised visual representation learning.
\newblock In \emph{Proceedings of the IEEE/CVF conference on computer vision and pattern recognition}, pp.\  9729--9738, 2020.

\bibitem[He et~al.(2022)He, Chen, Xie, Li, Doll{\'a}r, and Girshick]{he2022mae}
He, K., Chen, X., Xie, S., Li, Y., Doll{\'a}r, P., and Girshick, R.
\newblock Masked autoencoders are scalable vision learners.
\newblock In \emph{Proceedings of the IEEE/CVF conference on computer vision and pattern recognition}, pp.\  16000--16009, 2022.

\bibitem[Hessel et~al.(2018)Hessel, Modayil, Van~Hasselt, Schaul, Ostrovski, Dabney, Horgan, Piot, Azar, and Silver]{hessel2018rainbow}
Hessel, M., Modayil, J., Van~Hasselt, H., Schaul, T., Ostrovski, G., Dabney, W., Horgan, D., Piot, B., Azar, M., and Silver, D.
\newblock Rainbow: Combining improvements in deep reinforcement learning.
\newblock In \emph{Proc. the AAAI Conference on Artificial Intelligence (AAAI)}, 2018.

\bibitem[Hu et~al.(2023)Hu, Wang, Li, and Gao]{hu2023policybench}
Hu, Y., Wang, R., Li, L.~E., and Gao, Y.
\newblock For pre-trained vision models in motor control, not all policy learning methods are created equal.
\newblock \emph{arXiv preprint arXiv:2304.04591}, 2023.

\bibitem[Ioffe \& Szegedy(2015)Ioffe and Szegedy]{ioffe2015batch}
Ioffe, S. and Szegedy, C.
\newblock Batch normalization: Accelerating deep network training by reducing internal covariate shift.
\newblock In \emph{Proc. the International Conference on Machine Learning (ICML)}, 2015.

\bibitem[Islam et~al.(2022)Islam, Tomar, Lamb, Efroni, Zang, Didolkar, Misra, Li, van Seijen, Combes, et~al.]{islam2022acro}
Islam, R., Tomar, M., Lamb, A., Efroni, Y., Zang, H., Didolkar, A., Misra, D., Li, X., van Seijen, H., Combes, R. T.~d., et~al.
\newblock Agent-controller representations: Principled offline rl with rich exogenous information.
\newblock \emph{Proc. the International Conference on Machine Learning (ICML)}, 2022.

\bibitem[Kaiser et~al.(2019)Kaiser, Babaeizadeh, Mi{\l}os, Osi{\'n}ski, Campbell, Czechowski, Erhan, Finn, Kozakowski, Levine, et~al.]{kaiser2019model}
Kaiser, {\L}., Babaeizadeh, M., Mi{\l}os, P., Osi{\'n}ski, B., Campbell, R.~H., Czechowski, K., Erhan, D., Finn, C., Kozakowski, P., Levine, S., et~al.
\newblock Model based reinforcement learning for atari.
\newblock In \emph{Proc. the International Conference on Learning Representations (ICLR)}, 2019.

\bibitem[Kumar et~al.(2020)Kumar, Zhou, Tucker, and Levine]{kumar2020cql}
Kumar, A., Zhou, A., Tucker, G., and Levine, S.
\newblock Conservative q-learning for offline reinforcement learning.
\newblock \emph{Proc. the Advances in Neural Information Processing Systems (NeurIPS)}, 2020.

\bibitem[Kumar et~al.(2022{\natexlab{a}})Kumar, Agarwal, Geng, Tucker, and Levine]{kumar2022sql}
Kumar, A., Agarwal, R., Geng, X., Tucker, G., and Levine, S.
\newblock Offline q-learning on diverse multi-task data both scales and generalizes.
\newblock \emph{Proc. the International Conference on Learning Representations (ICLR)}, 2022{\natexlab{a}}.

\bibitem[Kumar et~al.(2022{\natexlab{b}})Kumar, Singh, Ebert, Yang, Finn, and Levine]{kumar2022ptr}
Kumar, A., Singh, A., Ebert, F., Yang, Y., Finn, C., and Levine, S.
\newblock Pre-training for robots: Offline rl enables learning new tasks from a handful of trials. arxiv e-prints, art.
\newblock \emph{arXiv preprint arXiv:2210.05178}, 2022{\natexlab{b}}.

\bibitem[Lan et~al.(2022)Lan, Tu, Oberman, Agarwal, and Bellemare]{lan2022generalization}
Lan, C.~L., Tu, S., Oberman, A., Agarwal, R., and Bellemare, M.~G.
\newblock On the generalization of representations in reinforcement learning.
\newblock \emph{arXiv preprint arXiv:2203.00543}, 2022.

\bibitem[Laskin et~al.(2020{\natexlab{a}})Laskin, Lee, Stooke, Pinto, Abbeel, and Srinivas]{laskin2020rad}
Laskin, M., Lee, K., Stooke, A., Pinto, L., Abbeel, P., and Srinivas, A.
\newblock Reinforcement learning with augmented data.
\newblock \emph{Proc. the Advances in Neural Information Processing Systems (NeurIPS)}, 2020{\natexlab{a}}.

\bibitem[Laskin et~al.(2020{\natexlab{b}})Laskin, Srinivas, and Abbeel]{laskin2020curl}
Laskin, M., Srinivas, A., and Abbeel, P.
\newblock Curl: Contrastive unsupervised representations for reinforcement learning.
\newblock In \emph{Proc. the International Conference on Machine Learning (ICML)}, 2020{\natexlab{b}}.

\bibitem[Lee et~al.(2023)Lee, Lee, Hwang, Lee, Lee, and Choo]{lee2023simtpr}
Lee, H., Lee, K., Hwang, D., Lee, H., Lee, B., and Choo, J.
\newblock On the importance of feature decorrelation for unsupervised representation learning in reinforcement learning.
\newblock \emph{arXiv preprint arXiv:2306.05637}, 2023.

\bibitem[Lee et~al.(2024{\natexlab{a}})Lee, Cho, Kim, Gwak, Kim, Choo, Yun, and Yun]{lee2024plastic}
Lee, H., Cho, H., Kim, H., Gwak, D., Kim, J., Choo, J., Yun, S.-Y., and Yun, C.
\newblock Plastic: Improving input and label plasticity for sample efficient reinforcement learning.
\newblock \emph{Advances in Neural Information Processing Systems}, 36, 2024{\natexlab{a}}.

\bibitem[Lee et~al.(2024{\natexlab{b}})Lee, Cho, Kim, Kim, Min, Choo, and Lyle]{lee2024hare}
Lee, H., Cho, H., Kim, H., Kim, D., Min, D., Choo, J., and Lyle, C.
\newblock Slow and steady wins the race: Maintaining plasticity with hare and tortoise networks.
\newblock \emph{arXiv e-prints}, pp.\  arXiv--2406, 2024{\natexlab{b}}.

\bibitem[Lee et~al.(2022{\natexlab{a}})Lee, Nachum, Yang, Lee, Freeman, Guadarrama, Fischer, Xu, Jang, Michalewski, et~al.]{lee2022multidt}
Lee, K.-H., Nachum, O., Yang, M.~S., Lee, L., Freeman, D., Guadarrama, S., Fischer, I., Xu, W., Jang, E., Michalewski, H., et~al.
\newblock Multi-game decision transformers.
\newblock \emph{Proc. the Advances in Neural Information Processing Systems (NeurIPS)}, 2022{\natexlab{a}}.

\bibitem[Lee et~al.(2022{\natexlab{b}})Lee, Nachum, Zhang, Guadarrama, Tan, and Yu]{lee2022piars}
Lee, K.-H., Nachum, O., Zhang, T., Guadarrama, S., Tan, J., and Yu, W.
\newblock Pi-ars: Accelerating evolution-learned visual-locomotion with predictive information representations.
\newblock In \emph{2022 IEEE/RSJ International Conference on Intelligent Robots and Systems (IROS)}, 2022{\natexlab{b}}.

\bibitem[Levine et~al.(2020)Levine, Kumar, Tucker, and Fu]{levine2020offline_rl_survey}
Levine, S., Kumar, A., Tucker, G., and Fu, J.
\newblock Offline reinforcement learning: Tutorial, review, and perspectives on open problems.
\newblock \emph{arXiv preprint arXiv:2005.01643}, 2020.

\bibitem[Lomonaco et~al.(2020)Lomonaco, Desai, Culurciello, and Maltoni]{lomonaco2020continual}
Lomonaco, V., Desai, K., Culurciello, E., and Maltoni, D.
\newblock Continual reinforcement learning in 3d non-stationary environments.
\newblock In \emph{Proceedings of the IEEE/CVF Conference on Computer Vision and Pattern Recognition Workshops}, pp.\  248--249, 2020.

\bibitem[Loshchilov \& Hutter(2017)Loshchilov and Hutter]{loshchilov2017decoupled}
Loshchilov, I. and Hutter, F.
\newblock Decoupled weight decay regularization.
\newblock \emph{Proc. the International Conference on Learning Representations (ICLR)}, 2017.

\bibitem[Ma et~al.(2022)Ma, Sodhani, Jayaraman, Bastani, Kumar, and Zhang]{ma2022vip}
Ma, Y.~J., Sodhani, S., Jayaraman, D., Bastani, O., Kumar, V., and Zhang, A.
\newblock Vip: Towards universal visual reward and representation via value-implicit pre-training.
\newblock 2022.

\bibitem[Machado et~al.(2018)Machado, Bellemare, Talvitie, Veness, Hausknecht, and Bowling]{machado2018atari}
Machado, M.~C., Bellemare, M.~G., Talvitie, E., Veness, J., Hausknecht, M., and Bowling, M.
\newblock Revisiting the arcade learning environment: Evaluation protocols and open problems for general agents.
\newblock \emph{Journal of Artificial Intelligence Research}, 2018.

\bibitem[Majumdar et~al.(2023{\natexlab{a}})Majumdar, Yadav, Arnaud, Ma, Chen, Silwal, Jain, Berges, Abbeel, Malik, et~al.]{majumdar2023cortexbench}
Majumdar, A., Yadav, K., Arnaud, S., Ma, Y.~J., Chen, C., Silwal, S., Jain, A., Berges, V.-P., Abbeel, P., Malik, J., et~al.
\newblock Where are we in the search for an artificial visual cortex for embodied intelligence?
\newblock 2023{\natexlab{a}}.

\bibitem[Majumdar et~al.(2023{\natexlab{b}})Majumdar, Yadav, Arnaud, Ma, Chen, Silwal, Jain, Berges, Abbeel, Malik, et~al.]{majumdar2023vc1}
Majumdar, A., Yadav, K., Arnaud, S., Ma, Y.~J., Chen, C., Silwal, S., Jain, A., Berges, V.-P., Abbeel, P., Malik, J., et~al.
\newblock Where are we in the search for an artificial visual cortex for embodied intelligence?
\newblock \emph{arXiv preprint arXiv:2303.18240}, 2023{\natexlab{b}}.

\bibitem[Mendonca et~al.(2023)Mendonca, Bahl, and Pathak]{mendonca2023structured}
Mendonca, R., Bahl, S., and Pathak, D.
\newblock Structured world models from human videos.
\newblock \emph{arXiv preprint arXiv:2308.10901}, 2023.

\bibitem[Mnih et~al.(2015)Mnih, Kavukcuoglu, Silver, Rusu, Veness, Bellemare, Graves, Riedmiller, Fidjeland, Ostrovski, et~al.]{mnih2015dqn}
Mnih, V., Kavukcuoglu, K., Silver, D., Rusu, A.~A., Veness, J., Bellemare, M.~G., Graves, A., Riedmiller, M., Fidjeland, A.~K., Ostrovski, G., et~al.
\newblock Human-level control through deep reinforcement learning.
\newblock \emph{Nature}, 2015.

\bibitem[Muhammad \& Yeasin(2020)Muhammad and Yeasin]{muhammad2020eigencam}
Muhammad, M.~B. and Yeasin, M.
\newblock Eigen-cam: Class activation map using principal components.
\newblock In \emph{2020 International Joint Conference on Neural Networks (IJCNN)}. IEEE, 2020.

\bibitem[Nair et~al.(2022)Nair, Rajeswaran, Kumar, Finn, and Gupta]{nair2022r3m}
Nair, S., Rajeswaran, A., Kumar, V., Finn, C., and Gupta, A.
\newblock R3m: A universal visual representation for robot manipulation.
\newblock \emph{6th Annual Conference on Robot Learning}, 2022.

\bibitem[Nakamoto et~al.(2023)Nakamoto, Zhai, Singh, Mark, Ma, Finn, Kumar, and Levine]{nakamoto2023calql}
Nakamoto, M., Zhai, Y., Singh, A., Mark, M.~S., Ma, Y., Finn, C., Kumar, A., and Levine, S.
\newblock Cal-ql: Calibrated offline rl pre-training for efficient online fine-tuning.
\newblock \emph{Proc. the Advances in Neural Information Processing Systems (NeurIPS)}, 2023.

\bibitem[Nikishin et~al.(2022)Nikishin, Schwarzer, D’Oro, Bacon, and Courville]{nikishin2022primacy}
Nikishin, E., Schwarzer, M., D’Oro, P., Bacon, P.-L., and Courville, A.
\newblock The primacy bias in deep reinforcement learning.
\newblock pp.\  16828--16847, 2022.

\bibitem[Padalkar et~al.(2023)Padalkar, Pooley, Jain, Bewley, Herzog, Irpan, Khazatsky, Rai, Singh, Brohan, et~al.]{padalkar2023rtx}
Padalkar, A., Pooley, A., Jain, A., Bewley, A., Herzog, A., Irpan, A., Khazatsky, A., Rai, A., Singh, A., Brohan, A., et~al.
\newblock Open x-embodiment: Robotic learning datasets and rt-x models.
\newblock \emph{arXiv preprint arXiv:2310.08864}, 2023.

\bibitem[Parisi et~al.(2022)Parisi, Rajeswaran, Purushwalkam, and Gupta]{parisi2022pvr}
Parisi, S., Rajeswaran, A., Purushwalkam, S., and Gupta, A.
\newblock The unsurprising effectiveness of pre-trained vision models for control.
\newblock In \emph{International Conference on Machine Learning}, pp.\  17359--17371. PMLR, 2022.

\bibitem[Pomerleau(1991)]{pomerleau1991bc}
Pomerleau, D.~A.
\newblock Efficient training of artificial neural networks for autonomous navigation.
\newblock \emph{Neural computation}, 3\penalty0 (1):\penalty0 88--97, 1991.

\bibitem[Radford et~al.(2021)Radford, Kim, Hallacy, Ramesh, Goh, Agarwal, Sastry, Askell, Mishkin, Clark, et~al.]{radford2021clip}
Radford, A., Kim, J.~W., Hallacy, C., Ramesh, A., Goh, G., Agarwal, S., Sastry, G., Askell, A., Mishkin, P., Clark, J., et~al.
\newblock Learning transferable visual models from natural language supervision.
\newblock In \emph{International conference on machine learning}, pp.\  8748--8763. PMLR, 2021.

\bibitem[Savva et~al.(2019)Savva, Kadian, Maksymets, Zhao, Wijmans, Jain, Straub, Liu, Koltun, Malik, et~al.]{savva2019habitat}
Savva, M., Kadian, A., Maksymets, O., Zhao, Y., Wijmans, E., Jain, B., Straub, J., Liu, J., Koltun, V., Malik, J., et~al.
\newblock Habitat: A platform for embodied ai research.
\newblock In \emph{Proceedings of the IEEE/CVF international conference on computer vision}, pp.\  9339--9347, 2019.

\bibitem[Schwarzer et~al.(2020{\natexlab{a}})Schwarzer, Anand, Goel, Hjelm, Courville, and Bachman]{schwarzer2020data}
Schwarzer, M., Anand, A., Goel, R., Hjelm, R.~D., Courville, A., and Bachman, P.
\newblock Data-efficient reinforcement learning with self-predictive representations.
\newblock \emph{Proc. the International Conference on Learning Representations (ICLR)}, 2020{\natexlab{a}}.

\bibitem[Schwarzer et~al.(2020{\natexlab{b}})Schwarzer, Anand, Goel, Hjelm, Courville, and Bachman]{schwarzer2020spr}
Schwarzer, M., Anand, A., Goel, R., Hjelm, R.~D., Courville, A., and Bachman, P.
\newblock Data-efficient reinforcement learning with self-predictive representations.
\newblock In \emph{Proc. the International Conference on Learning Representations (ICLR)}, 2020{\natexlab{b}}.

\bibitem[Schwarzer et~al.(2021{\natexlab{a}})Schwarzer, Rajkumar, Noukhovitch, Anand, Charlin, Hjelm, Bachman, and Courville]{schwarzer2021pretraining}
Schwarzer, M., Rajkumar, N., Noukhovitch, M., Anand, A., Charlin, L., Hjelm, R.~D., Bachman, P., and Courville, A.~C.
\newblock Pretraining representations for data-efficient reinforcement learning.
\newblock \emph{Proc. the Advances in Neural Information Processing Systems (NeurIPS)}, 2021{\natexlab{a}}.

\bibitem[Schwarzer et~al.(2021{\natexlab{b}})Schwarzer, Rajkumar, Noukhovitch, Anand, Charlin, Hjelm, Bachman, and Courville]{schwarzer2021sgi}
Schwarzer, M., Rajkumar, N., Noukhovitch, M., Anand, A., Charlin, L., Hjelm, R.~D., Bachman, P., and Courville, A.~C.
\newblock Pretraining representations for data-efficient reinforcement learning.
\newblock \emph{Proc. the Advances in Neural Information Processing Systems (NeurIPS)}, 2021{\natexlab{b}}.

\bibitem[Schwarzer et~al.(2023)Schwarzer, Ceron, Courville, Bellemare, Agarwal, and Castro]{schwarzer2023bbf}
Schwarzer, M., Ceron, J. S.~O., Courville, A., Bellemare, M.~G., Agarwal, R., and Castro, P.~S.
\newblock Bigger, better, faster: Human-level atari with human-level efficiency.
\newblock In \emph{International Conference on Machine Learning}, pp.\  30365--30380. PMLR, 2023.

\bibitem[Seo et~al.(2022)Seo, Lee, James, and Abbeel]{seo2022video-ptr}
Seo, Y., Lee, K., James, S.~L., and Abbeel, P.
\newblock Reinforcement learning with action-free pre-training from videos.
\newblock In \emph{International Conference on Machine Learning}, pp.\  19561--19579. PMLR, 2022.

\bibitem[Seo et~al.(2023{\natexlab{a}})Seo, Hafner, Liu, Liu, James, Lee, and Abbeel]{seo2023mwm}
Seo, Y., Hafner, D., Liu, H., Liu, F., James, S., Lee, K., and Abbeel, P.
\newblock Masked world models for visual control.
\newblock In \emph{6th Annual Conference on Robot Learning}, pp.\  1332--1344. PMLR, 2023{\natexlab{a}}.

\bibitem[Seo et~al.(2023{\natexlab{b}})Seo, Kim, James, Lee, Shin, and Abbeel]{seo2023mvmwm}
Seo, Y., Kim, J., James, S., Lee, K., Shin, J., and Abbeel, P.
\newblock Multi-view masked world models for visual robotic manipulation.
\newblock \emph{Proc. the International Conference on Machine Learning (ICML)}, 2023{\natexlab{b}}.

\bibitem[Sermanet et~al.(2018)Sermanet, Lynch, Chebotar, Hsu, Jang, Schaal, Levine, and Brain]{sermanet2018time_contrast}
Sermanet, P., Lynch, C., Chebotar, Y., Hsu, J., Jang, E., Schaal, S., Levine, S., and Brain, G.
\newblock Time-contrastive networks: Self-supervised learning from video.
\newblock In \emph{2018 IEEE international conference on robotics and automation (ICRA)}, pp.\  1134--1141. IEEE, 2018.

\bibitem[Shah et~al.(2023)Shah, Mart{\'\i}n-Mart{\'\i}n, and Zhu]{shah2023mutex}
Shah, R., Mart{\'\i}n-Mart{\'\i}n, R., and Zhu, Y.
\newblock Mutex: Learning unified policies from multimodal task specifications.
\newblock \emph{arXiv preprint arXiv:2309.14320}, 2023.

\bibitem[Stone et~al.(2021)Stone, Ramirez, Konolige, and Jonschkowski]{stone2021dcs}
Stone, A., Ramirez, O., Konolige, K., and Jonschkowski, R.
\newblock The distracting control suite--a challenging benchmark for reinforcement learning from pixels.
\newblock \emph{arXiv preprint arXiv:2101.02722}, 2021.

\bibitem[Stooke et~al.(2021)Stooke, Lee, Abbeel, and Laskin]{stooke2021atc}
Stooke, A., Lee, K., Abbeel, P., and Laskin, M.
\newblock Decoupling representation learning from reinforcement learning.
\newblock In \emph{Proc. the International Conference on Machine Learning (ICML)}, 2021.

\bibitem[Taiga et~al.(2022)Taiga, Agarwal, Farebrother, Courville, and Bellemare]{taiga2022investigating}
Taiga, A.~A., Agarwal, R., Farebrother, J., Courville, A., and Bellemare, M.~G.
\newblock Investigating multi-task pretraining and generalization in reinforcement learning.
\newblock In \emph{Proc. the International Conference on Learning Representations (ICLR)}, 2022.

\bibitem[Tassa et~al.(2018)Tassa, Doron, Muldal, Erez, Li, Casas, Budden, Abdolmaleki, Merel, Lefrancq, et~al.]{tassa2018dmc}
Tassa, Y., Doron, Y., Muldal, A., Erez, T., Li, Y., Casas, D. d.~L., Budden, D., Abdolmaleki, A., Merel, J., Lefrancq, A., et~al.
\newblock Deepmind control suite.
\newblock \emph{arXiv preprint arXiv:1801.00690}, 2018.

\bibitem[Tong et~al.(2022)Tong, Song, Wang, and Wang]{tong2022videomae}
Tong, Z., Song, Y., Wang, J., and Wang, L.
\newblock Videomae: Masked autoencoders are data-efficient learners for self-supervised video pre-training.
\newblock \emph{Advances in neural information processing systems}, 35:\penalty0 10078--10093, 2022.

\bibitem[Walke et~al.(2023)Walke, Black, Zhao, Vuong, Zheng, Hansen-Estruch, He, Myers, Kim, Du, et~al.]{walke2023bridgedata}
Walke, H.~R., Black, K., Zhao, T.~Z., Vuong, Q., Zheng, C., Hansen-Estruch, P., He, A.~W., Myers, V., Kim, M.~J., Du, M., et~al.
\newblock Bridgedata v2: A dataset for robot learning at scale.
\newblock In \emph{Conference on Robot Learning}, pp.\  1723--1736. PMLR, 2023.

\bibitem[Wang et~al.(2022)Wang, Luo, Ross, and Li]{wang2022vrl3}
Wang, C., Luo, X., Ross, K., and Li, D.
\newblock Vrl3: A data-driven framework for visual deep reinforcement learning.
\newblock \emph{Advances in Neural Information Processing Systems}, 35:\penalty0 32974--32988, 2022.

\bibitem[Wu et~al.(2023)Wu, Majumdar, Stone, Lin, Mordatch, Abbeel, and Rajeswaran]{wu2023maskedfd}
Wu, P., Majumdar, A., Stone, K., Lin, Y., Mordatch, I., Abbeel, P., and Rajeswaran, A.
\newblock Masked trajectory models for prediction, representation, and control.
\newblock \emph{arXiv preprint arXiv:2305.02968}, 2023.

\bibitem[Wu \& He(2018)Wu and He]{wu2018group}
Wu, Y. and He, K.
\newblock Group normalization.
\newblock In \emph{Proc. of the European Conference on Computer Vision (ECCV)}, pp.\  3--19, 2018.

\bibitem[Xiao et~al.(2021)Xiao, Singh, Mintun, Darrell, Doll{\'a}r, and Girshick]{xiao2021early}
Xiao, T., Singh, M., Mintun, E., Darrell, T., Doll{\'a}r, P., and Girshick, R.
\newblock Early convolutions help transformers see better.
\newblock \emph{Advances in neural information processing systems}, 34:\penalty0 30392--30400, 2021.

\bibitem[Xiao et~al.(2022)Xiao, Radosavovic, Darrell, and Malik]{xiao2022maskedvismotor}
Xiao, T., Radosavovic, I., Darrell, T., and Malik, J.
\newblock Masked visual pre-training for motor control.
\newblock \emph{arXiv preprint arXiv:2203.06173}, 2022.

\bibitem[Yang et~al.(2023)Yang, Nachum, Du, Wei, Abbeel, and Schuurmans]{yang2023foundation_survey}
Yang, S., Nachum, O., Du, Y., Wei, J., Abbeel, P., and Schuurmans, D.
\newblock Foundation models for decision making: Problems, methods, and opportunities.
\newblock \emph{arXiv preprint arXiv:2303.04129}, 2023.

\bibitem[Ye et~al.(2022)Ye, Zhang, Abbeel, and Gao]{ye2022ficc}
Ye, W., Zhang, Y., Abbeel, P., and Gao, Y.
\newblock Become a proficient player with limited data through watching pure videos.
\newblock In \emph{The Eleventh International Conference on Learning Representations}, 2022.

\bibitem[Yu et~al.(2021)Yu, Lan, Zeng, Feng, Zhang, and Chen]{yu2021playvirtual}
Yu, T., Lan, C., Zeng, W., Feng, M., Zhang, Z., and Chen, Z.
\newblock Playvirtual: Augmenting cycle-consistent virtual trajectories for reinforcement learning.
\newblock \emph{Advances in Neural Information Processing Systems}, 34:\penalty0 5276--5289, 2021.

\bibitem[Yu et~al.(2022)Yu, Zhang, Lan, Chen, and Lu]{yu2022mlr}
Yu, T., Zhang, Z., Lan, C., Chen, Z., and Lu, Y.
\newblock Mask-based latent reconstruction for reinforcement learning.
\newblock \emph{Proc. the Advances in Neural Information Processing Systems (NeurIPS)}, 2022.

\bibitem[Yuan et~al.(2023)Yuan, Yang, Hua, Chang, Hu, Wang, and Xu]{yuan2023rlvigen}
Yuan, Z., Yang, S., Hua, P., Chang, C., Hu, K., Wang, X., and Xu, H.
\newblock Rl-vigen: A reinforcement learning benchmark for visual generalization.
\newblock \emph{Proc. the Advances in Neural Information Processing Systems (NeurIPS)}, 2023.

\bibitem[Zang et~al.(2022)Zang, Li, Yu, Liu, Islam, Combes, and Laroche]{zang2022priorbc}
Zang, H., Li, X., Yu, J., Liu, C., Islam, R., Combes, R. T.~D., and Laroche, R.
\newblock Behavior prior representation learning for offline reinforcement learning.
\newblock \emph{arXiv preprint arXiv:2211.00863}, 2022.

\bibitem[Zhang et~al.(2022)Zhang, GX-Chen, Sobal, LeCun, and Carion]{zhang2022lightweight_probe}
Zhang, W., GX-Chen, A., Sobal, V., LeCun, Y., and Carion, N.
\newblock Light-weight probing of unsupervised representations for reinforcement learning.
\newblock \emph{arXiv preprint arXiv:2208.12345}, 2022.

\bibitem[Zheng et~al.(2023)Zheng, Wang, Sun, Ma, Zhao, Xu, Daum{\'e}~III, and Huang]{zheng2023taco}
Zheng, R., Wang, X., Sun, Y., Ma, S., Zhao, J., Xu, H., Daum{\'e}~III, H., and Huang, F.
\newblock Taco: Temporal latent action-driven contrastive loss for visual reinforcement learning.
\newblock \emph{arXiv preprint arXiv:2306.13229}, 2023.

\bibitem[Zhu et~al.(2020)Zhu, Wong, Mandlekar, Mart{\'\i}n-Mart{\'\i}n, Joshi, Nasiriany, and Zhu]{zhu2020robosuite}
Zhu, Y., Wong, J., Mandlekar, A., Mart{\'\i}n-Mart{\'\i}n, R., Joshi, A., Nasiriany, S., and Zhu, Y.
\newblock robosuite: A modular simulation framework and benchmark for robot learning.
\newblock \emph{arXiv preprint arXiv:2009.12293}, 2020.

\end{thebibliography}
\bibliographystyle{icml2024}

\newpage
\appendix
\onecolumn



\section{Extended Related Work}

The availability of large-scale offline datasets tailored for robotics \citep{walke2023bridgedata, shah2023mutex, mendonca2023structured} has significantly influenced reinforcement learning (RL) research. 
This influence is particularly notable in adopting a pre-trained visual encoder for various downstream tasks \citep{ye2022ficc, ma2022vip, lee2023simtpr}.

A crucial aspect in the selection of pre-trained models is the type of data employed during pre-training. These algorithms are typically categorized by the data they use: Images, Videos, Demonstrations, and Trajectories. Each category utilizes different objectives to extract distinct features during the pre-training phase.

\textbf{Image:} 
Image-based learning encompasses two main strategies. The first focuses on augmentation-invariant representations, achieved by ensuring consistency in the latent space for different augmentations of the same image \citep{chen2020simclr, laskin2020curl, chen2021simsiam, chen2020mocov2, grill2020byol}. The second strategy involves reconstructing heavily masked images, leveraging transformer architectures \citep{he2022mae, seo2023mwm, seo2023mvmwm}.

\textbf{Video:} 
Video-based algorithms extend image-based techniques by integrating temporality. 
Temporal contrastive learning, for example, aims to closely encode temporally adjacent images to understand temporal dynamics \citep{nair2022r3m, ma2022vip, sermanet2018time_contrast, han2019dense_cpc, stooke2021atc}.  Other approaches include reconstructing masked images from temporally adjacent frames \citep{yu2022mlr, gupta2023siammae, tong2022videomae, feichtenhofer2022mae-st}, or autoregressively predicting future frames based on past observations \citep{hafner2019rssm, seo2022video-ptr}.

\textbf{Demonstration: } 
Learning from demonstrations has a wide range of methods. 
Inverse dynamics learning focuses on predicting actions, given consecutive states \citep{christiano2016idm, islam2022acro, brandfonbrener2023idm-theory}. Forward dynamics learning targets predicting future states from current state-action pairs \citep{schwarzer2020spr, yu2021playvirtual, lee2023simtpr, zheng2023taco}. Imitation learning, on the other hand, aims to replicate the behavior policy demonstrated in the data \citep{pomerleau1991bc, zang2022priorbc, arora2020provablebc, baker2022vpt, caluwaerts2023barkour}. Some approaches combine multiple methods for enhanced robustness \citet{yu2022mlr, zhang2022lightweight_probe}.

\textbf{Trajectory:} 
Trajectory-based methods maximize the use of reward information available in trajectories. These can be broadly categorized into two settings: online, where data is collected through real-time interaction with the environment \citep{mnih2015dqn, bellemare2017c51, hafner2019dreamer, laskin2020rad}, and offline, which involves learning from pre-existing datasets \citep{fujimoto2019bcq, kumar2020cql, lee2022multidt, nakamoto2023calql, wu2023maskedfd, lee2022piars}.




Recent studies have evaluated the effectiveness of pre-trained visual representations in vision-based RL, employing a variety of pre-training methods \cite{parisi2022pvr, majumdar2023vc1, hu2023policybench}. These investigations have used different models, including ResNet \cite{he2016resnet} and ViT \cite{dosovitskiy2020vit}, which were pre-trained on a wide array of datasets \cite{deng2009imagenet, savva2019habitat, tassa2018dmc, grauman2022ego4d}. The findings indicate that while visual representations pre-trained on large and diverse datasets can enhance generalization in downstream tasks, no single pre-trained model consistently excels in generalization across all types of tasks.


\newpage

\section{Implementation Details for Atari-PB}
\label{appendix:impl_details}

\subsection{Pre-training}
\label{appendix:impl_pretrain}

In this section, we describe our pre-training stage in detail.
We first depict the three components of our pre-training model (backbone, neck, and head) and explain how our pre-training dataset was curated. Note that the model description is our most fundamental form, and that adjustments are often made to align with each method's requirements. For such algorithm-specific elements, refer to Section \ref{appendix:baselines}.

\textbf{Backbone, $f(\cdot)$:} A backbone network is a game-agnostic spatial feature extractor. We employ a widely used and sufficiently large ResNet-50, but replace batch normalization with group normalization following \citet{kumar2022sql}. Given an $(4,84,84)$ input $\mathbf{o}$, the backbone encodes them into a $(2048,6,6)$ feature map $\mathbf{z}=f(\mathbf{o})$.

\textbf{Neck, $g(\cdot)$:} The main goal of neck is to encode the feature map into a $512$-dimensional vector $\mathbf{q}=g(\mathbf{z})$, while applying learnable spatial embedding. Given the backbone output, each feature map is point-wise multiplied with its game-specific spatial embedding. Spatial pooling and instance normalization is then applied to obtain a $2048$-dimensional vector, which is further encoded by a neural network. Unless stated otherwise, we use a 2-layer MLP with ReLU activation, hidden dimension of $1024$, and output dimension of $512$.

\textbf{Head, $h(\cdot)$:} The head makes the prediction $\mathbf{y}=h(\mathbf{q})$, where $\mathbf{y}$'s dimensionality depends on the task. We employ a multi-head architecture, meaning that multiple neural networks of identical architecture are trained, each one devoted for each game. Unless specified, we use a single linear layer that outputs the action prediction $\mathbf{y}\in \mathbb{R}^{A}$.


\textbf{Dataset:}
The DQN-Replay-Dataset \citep{agarwal2020optimistic}, a collection of DQN agent's training logs in 60 Atari games, provides 50 million transitions for each game collected across five different runs. These runs are subdivided into 50 checkpoints, each containing 1 million transitions of differing optimality. To fulfill our desiderata of "reflecting the diverse nature of real-world datasets" while keeping it computationally accessible, we choose to compile small segments from multiple runs and checkpoints. Our data creation procedure is thus the following: from the 50 games of our choice, we choose the first 2 runs of each game and the first 10 checkpoints of each run. From each of the 1,000 checkpoints, the initial 10,000 interactions are sampled, resulting in a 10 million dataset. We found the initial 10 checkpoints to be sufficient for covering both suboptimal and expert policies; as shown by the supplementary figures of \citet{agarwal2020optimistic}, 40 million steps (end of 10th checkpoint) is enough for DQN agents to achieve reasonable score in all games.

\begin{table}[h]
\vspace{-3mm}
\begin{center}
\caption{Games categorized by distribution.}
\label{table:game_list}
\vspace{0.8ex}
\small{
\begin{tabular}{l p{10cm}}
\toprule
Distribution & Games \\
\midrule \\[-2.7ex]
In-Distribution & AirRaid, Amidar, Asteroids, Atlantis, BankHeist, \newline
                  BattleZone, Berzerk, Bowling, Boxing, Breakout, \newline
                  Carnival, Centipede, ChopperCommand, CrazyClimber, DemonAttack, \newline
                  DoubleDunk, ElevatorAction, Enduro, FishingDerby, Freeway, \newline
                  Frostbite, Gopher, Gravitar, Hero, IceHockey, \newline
                  Jamesbond, Kangaroo, Krull, KungFuMaster, MontezumaRevenge, \newline
                  MsPacman, NameThisGame, Phoenix, Pitfall, PrivateEye, \newline
                  Qbert, RoadRunner, Robotank, Skiing, Solaris, \newline
                  SpaceInvaders, StarGunner, Tennis, TimePilot, Tutankham, \newline
                  UpNDown, VideoPinball, WizardOfWor, YarsRevenge, Zaxxon \\
\midrule
Near-Out-of-Distribution & Alien, Assault, Asterix, BeamRider, JourneyEscape, \newline
                                    Pong, Pooyan, Riverraid, Seaquest, Venture \\
\midrule
Far-Out-of-Distribution & BasicMath, HumanCannonball, Klax, Othello, Surround\\
\bottomrule
\end{tabular}}
\end{center}
\end{table}

\newpage

\subsection{Baseline Implementations}
\label{appendix:baselines}

We provide a more detailed description of each method in the following sections. Table \ref{table:pretrain_hyperparameter} shows the universal hyperparameters across all pre-training methods; any method-specific hyperparameters are individually listed in the following tables.

\begin{table}[h]
\vspace{-3mm}
\begin{center}
\caption{Global pre-training hyperparameters.}
\vspace{2mm}
\small
{
\begin{tabular}{lr}
\toprule
Hyperparameter & Value \\
\midrule \\[-2.5ex]
Observation rendering    & (84,84), Grayscale \\
Frames stacked           & 4 \\
Reward discount factor ($\gamma$) & 1.0 (DT) \\
                         & 0.99 (Rest) \\
Action space size ($|A|$)       & 18 \\
\midrule \\[-2.5ex]
Augmentation             & [Random Shift, Intensity] \\
Random shift pad         & 4 \\
Intensity scale          & 0.05 \\
\midrule \\[-2.5ex]
Learning rate scheduler  & Cosine annealing with warmup \\
Warmup ratio             & 0.1 \\
Initial learning rate ratio & 0.1 \\
\bottomrule
\end{tabular}}
\label{table:pretrain_hyperparameter}
\end{center}
\vspace{-3mm}
\end{table}

\subsubsection{CURL}
Contrastive Unsupervised Representations for Reinforcement Learning \citep{laskin2020curl} learns augmentation invariant representations using InfoNCE loss and momentum encoder. Given two augmented versions (two views) of an image (denoted $\mathbf{o}_1,\mathbf{o}_2$), one is passed through an 'online' encoder to get $\mathbf{q}=g(f(\mathbf{o}_1))$ and the other is passed through a coupled 'momentum' encoder to get a target $\mathbf{q}_+=g'(f'(\mathbf{o}_2))$. To predict the target, $\mathbf{q}$ is passed through a predictor network to get $\mathbf{y}=h(\mathbf{q})$. 
The InfoNCE loss is then computed based on $\mathbf{y}$ and $\mathbf{q}_+$:
$$\mathcal{L}_{\text{CURL}} = -\sum_{b\in B}\log \frac{\exp(\mathbf{y}^b \cdot \mathbf{q}^b_+)}{\exp(\mathbf{y}^b \cdot \mathbf{q}^b_+) + \sum_{b'\in B-\{b\}}\exp(\mathbf{y}^b \cdot \mathbf{q}^{b'}_+)}$$
As implied in the notations, we use backbone and neck as the encoder and head as the predictor. The momentum networks $f',g'$ are updated every iteration with a coefficient of $\tau = 0.99$ that scales linearly up to $0.999$. As for the similarity measure, we use dot product instead of bilinear product \citep{chen2020mocov2}.

\begin{table}[h]
\vspace{-3mm}
\begin{center}
\caption{Hyperparameters for pre-training CURL.}
\vspace{2mm}
\small{
\begin{tabular}{lr}
\toprule
Hyperparameter & Value \\
\midrule \\[-2.5ex]
Epochs (early stop)      & 100 (20) \\
Base learning rate       & 3e-6 \\
Weight decay             & 1e-5 \\
Optimizer ($\beta_1$, $\beta_2$)
                         & AdamW (0.9, 0.999) \\
Batch size               & 512 \\
\midrule \\[-2.5ex]
Momentum update ratio ($\tau$)
                         & [0.9, 0.999] \\
Representation dimensions
                         & 512 \\
Similarity measure       & Dot product \\
\bottomrule
\end{tabular}}
\label{table:curl_hyperparameter}
\end{center}
\vspace{-3mm}
\end{table}

\subsubsection{MAE}
\label{section:mae_detail}
Masked Autoencoder \citep{he2022mae} learns to reconstruct heavily masked images with transformer encoder-decoder architecture. As we use a convolutional network for our backbone, we naturally turn from masking patches of images to masking pixels of convolutional features, inspired by \citet{seo2023mwm}. The transformer encoder-decoder module is added at the end of our neck, and the head is used to predict the target pixels.
In the neck, we perform masking after applying the game-wise spatial embedding, and pass the $2048$-dimensional features through a 2 layer MLP into a sequence of $512$-dimensional tokens. These are then processed with a transformer encoder, appended with mask tokens for prediction, and further processed with a transformer decoder. Finally, the head's game-wise linear layer is used to predict the pixels of the target image. Because the backbone reduces an $(84,84)$ image into a $(6,6)$ spatial embedding (or $36$ tokens), the game-wise linear layer maps each token to a $784$-dimensional vector to predict a $(4,14,14)$ sized patch of the target image.

We follow the hyperparameters provided by \citet{he2022mae} but use a higher mask ratio $\rho=0.9$. We use a 3-layer transformer encoder and a 4-layer transformer decoder, following \citet{seo2023mwm}.

\begin{table}[h]
\vspace{-3mm}
\begin{center}
\caption{Hyperparameters for pre-training MAE.}
\vspace{2mm}
\small{
\begin{tabular}{lr}
\toprule
Hyperparameter & Value \\
\midrule \\[-2.5ex]
Epochs                       & 100 \\
Base learning rate           & 3e-4 \\
Weight decay                 & 5e-2 \\
Optimizer ($\beta_1$, $\beta_2$)
                             & AdamW (0.9, 0.95) \\
Batch size                   & 512 \\
\midrule \\[-2.5ex]
Mask ratio ($\rho$)          & 0.9 \\
Transformer embedding dimensions
                             & 512 \\
Transformer MLP ratio        & 4 \\
Transformer heads            & 4 \\
Transformer encoder layers   & 3 \\
Transformer decoder layers   & 4 \\
(Head) Linear output dimensions       & 768 \\
\bottomrule
\end{tabular}}
\label{table:mae_hyperparameter}
\end{center}
\vspace{-3mm}
\end{table}

\subsubsection{ATC}
Augmented Temporal Contrast \citep{stooke2021atc} learns temporally predictive representations by maximizing the similarity between the representations of current states and their paired future states, using InfoNCE loss and momentum encoder. This means that the input image $\mathbf{o}_t$ is encoded by an 'online' encoder into $\mathbf{q}_t = g(f(\mathbf{o}_t))$, and its near-future target image $\mathbf{o}_{t+k}$ is encoded by a 'momentum' encoder into $\mathbf{q}_{t+k} = g'(f'(\mathbf{o}_{t+k}))$. 
With the two representations in hand, the predictor predicts the target $\mathbf{q}_{t+k}$ via $\mathbf{y}_t=h(\mathbf{q}_t)$ to minimize the InfoNCE loss:
$$\mathcal{L}_{\text{ATC}} = -\sum_{b\in B}\log \frac{\exp(\mathbf{y}^b_t \cdot \mathbf{q}^b_{t+k})}{\exp(\mathbf{y}^b_{t} \cdot \mathbf{q}^b_{t+k}) + \sum_{b'\in B-\{b\}}\exp(\mathbf{y}^b_{t} \cdot \mathbf{q}^{b'}_{t+k})}$$
Same as CURL, we use the backbone and neck as our encoder and the head as our predictor. The momentum update is performed every iteration with a coefficient of $\tau = 0.99$ that scales linearly up to $0.999$. We also use dot product instead of bilinear product as a similarity measure.

\begin{table}[h]
\vspace{-3mm}
\begin{center}
\caption{Hyperparameters for pre-training ATC.}
\vspace{2mm}
\small{
\begin{tabular}{lr}
\toprule
Hyperparameter & Value \\
\midrule \\[-2.5ex]
Epochs                   & 100 \\
Base learning rate       & 3e-4 \\
Weight decay             & 1e-5 \\
Optimizer ($\beta_1$, $\beta_2$)
                         & AdamW (0.9, 0.999) \\
Batch size               & 512 \\
\midrule \\[-2.5ex]
Steps to future state ($k$) & 3 \\
Momentum update ratio ($\tau$)
                         & [0.9, 0.999] \\
Representation dimensions
                         & 512 \\
Similarity measure       & Dot product \\
\bottomrule
\end{tabular}}
\label{table:atc_hyperparameter}
\end{center}
\vspace{-3mm}
\end{table}

\subsubsection{SiamMAE}
Siamese Masked Autoencoder \cite{gupta2023siammae} extends MAE \citep{he2022mae} to a temporal prediction task using siamese architecture and asymmetric masking strategy. Concretely, both current image $\mathbf{o}_t$ and future(target) image $\mathbf{o}_{t+k}$ are passed through the same network up until the transformer decoder. In our implementation, this includes the backbone, game-wise spatial embedding, and the transformer encoder. In the process, $\mathbf{o}_{t+k}$ is masked with an extremely high ratio($\rho=0.95$) while $\mathbf{o}_t$ remains untouched. In order to retrieve the lost information, the transformer decoder refers to the current frame's information via cross attention. For image reconstruction, we use the same head as our MAE. Overall, we follow the training procedure stated by \citet{gupta2023siammae} with some modifications. First, for the same reason as MAE, we use convolution feature masking instead of patch masking. Second, we use the transformer architecture used to train our MAE. Third, we reduce the future sampling window from $[4,48]$ to $[1,3]$.

\begin{table}[h]
\vspace{-3mm}
\begin{center}
\caption{Hyperparameters for pre-training SiamMAE.}
\vspace{2mm}
\small{
\begin{tabular}{lr}
\toprule
Hyperparameter & Value \\
\midrule \\[-2.5ex]
Epochs                   & 100 \\
Base learning rate       & 3e-4 \\
Weight decay             & 5e-2 \\
Optimizer ($\beta_1$, $\beta_2$)
                         & AdamW (0.9, 0.95) \\
Batch size               & 512 \\
\midrule \\[-2.5ex]
Steps to future state ($k$) & [1,3] \\
Mask ratio ($\rho$)          & 0.95 \\
Transformer embedding dimensions
                             & 512 \\
Transformer MLP ratio        & 4 \\
Transformer heads          & 4 \\
Transformer encoder layers & 3 \\
Transformer decoder layers & 4 \\
(Head) Linear output dimensions       & 768 \\
\bottomrule
\end{tabular}}
\label{table:siammae_hyperparameter}
\end{center}
\vspace{-3mm}
\end{table}

\subsubsection{R3M$^\dagger$}
Reusable Representations for Robot Manipulation \cite{nair2022r3m} uses multiple losses to learn from human demonstration videos with diverse tasks, but we focus our study on the time contrastive loss. Although similar to InfoNCE loss in ATC \citep{stooke2021atc}, the main difference is that in addition to the future(target) image $\mathbf{o}_{t+k}$, a further-future image $\mathbf{o}_{t+k'} (k<k')$ is sampled as a 'hard' negative. Unfortunately, we find pre-training with the original time contrastive loss to be challenging. In order to preserve the aforementioned idea, we implement R3M as 'ATC with an additional hard negative' and use the following loss function:
$$\mathcal{L}_{\text{R3M}} = -\sum_{b\in B}\log
\frac{\exp(\mathbf{y}^b_t \cdot \mathbf{q}^b_{t+k})}
{\exp(\mathbf{y}^b_{t} \cdot \mathbf{q}^b_{t+k})
    + \sum_{b'\in B-\{b\}}\exp(\mathbf{y}^b_{t} \cdot \mathbf{q}^{b'}_{t+k})
    + \exp(\mathbf{y}^b_{t} \cdot \mathbf{q}^b_{t+k'})
}
$$
Naturally, we use the same momentum networks as ATC; all future images are passed through momentum backbone and neck, which are updated every iteration.





\begin{table}[h]
\vspace{-3mm}
\begin{center}
\caption{Hyperparameters for pre-training R3M$^\dagger$.}
\vspace{2mm}
\small{
\begin{tabular}{lr}
\toprule
Hyperparameter & Value \\
\midrule \\[-2.5ex]
Epochs                   & 100 \\
Base learning rate       & 3e-4 \\
Weight decay             & 1e-5 \\
Optimizer ($\beta_1$, $\beta_2$)
                         & AdamW (0.9, 0.999) \\
Batch size               & 512 \\
\midrule \\[-2.5ex]
Steps to future states ($k, k'$) & 3, 6 \\
Momentum update ratio ($\tau$)
                         & [0.9, 0.999] \\
Output representation dimensions
                         & 512 \\
Similarity measure       & Dot product \\
\bottomrule
\end{tabular}}
\label{table:r3m_hyperparameter}
\end{center}
\vspace{-3mm}
\end{table}

\newpage

\subsubsection{BC} 
Behavioral cloning learns to predict the action $\mathbf{a}_t$ from its observation $\mathbf{o}_t$. Predictions are made by a single pass through the model $\mathbf{y}_t=h(g(f(\mathbf{o}_t)))$, of which goal is to minimize the cross entropy loss between $\mathbf{y}_t$ and $\mathbf{a}_t$.

\begin{table}[h]
\vspace{-3mm}
\begin{center}
\caption{Hyperparameters for pre-training BC.}
\vspace{2mm}
\resizebox{0.35 \textwidth}{!}
{
\begin{tabular}{lr}
\toprule
Hyperparameter & Value \\
\midrule \\[-2.5ex]
Epochs                   & 100 \\
Base learning rate       & 3e-4 \\
Weight decay             & 1e-5 \\
Optimizer ($\beta_1$, $\beta_2$)
                         & AdamW (0.9, 0.999) \\
Batch size               & 512 \\
\bottomrule
\end{tabular}}
\label{table:bc_hyperparameter}
\end{center}
\vspace{-3mm}
\end{table}

\subsubsection{SPR}
Self Predictive Representations \citep{schwarzer2020spr} learn to recursively predict future states from a starting state and subsequent actions. Given an image $\mathbf{o}_t$ and a sequence of actions $\mathbf{a}_{t:t+K-1}$, the model's goal is to predict the consequent future images $\mathbf{o}_{t+1:t+K}$ in the latent space. The current image is encoded using an 'online' network to get $\mathbf{q}_t = f(g(\mathbf{o}_t))$, which is used in tandem with the actions to predict future representations $\mathbf{y}_{t+1:t+K} = h(\mathbf{q}_t, \mathbf{a}_{t:t+K-1})$. The target representations are obtained by encoding the future images with a 'momentum' encoder: $\mathbf{q}_{i} = f'(g'(\mathbf{o}_{i})), i\in[t+1,t+K]$. In our experiments, we make the following modifications. First, we use RNN instead of recursive CNN, which is put in front of the linear layer in the head network. A game-wise action embedding for the RNN input, as each game uses a different set of actions. Second, we use contrastive loss instead of cosine similarity loss. Third, we discard the Q-learning loss. As a result, we use the following loss:
$$\mathcal{L}_{\text{SPR}} = -\sum_{b\in B} \sum^{K}_{k=1}\log
\frac{\exp(\mathbf{y}^b_{t+k} \cdot \mathbf{q}^b_{t+k})}
{ \sum_{b'\in B} \sum^{K}_{k'=1} \exp(\mathbf{y}^{b}_{t+k} \cdot \mathbf{q}^{b'}_{t+k'}) }
$$

\begin{table}[!h]
\vspace{-3mm}
\begin{center}
\caption{Hyperparameters for pre-training SPR.}
\vspace{2mm}
\small{
\begin{tabular}{lr}
\toprule
Hyperparameter & Value \\
\midrule \\[-2.5ex]
Epochs                   & 25 \\
Base learning rate       & 3e-4 \\
Weight decay             & 1e-4 \\
Optimizer ($\beta_1$, $\beta_2$)
                         & AdamW (0.9, 0.999) \\
Batch size               & 128 \\
\midrule \\[-2.5ex]
Prediction sequence length ($K$) & 4 \\
Momentum update ratio ($\tau$)
                         & [0.9, 0.999] \\
Representation dimensions
                         & 512 \\
Similarity measure       & Dot product \\
\bottomrule
\end{tabular}}
\label{table:spr_hyperparameter}
\end{center}
\vspace{-3mm}
\end{table}

\subsubsection{IDM}
Inverse Dynamics Modeling \cite{christiano2016idm} learns to predict the action $\mathbf{a}_t$ taken between two successive observations $\mathbf{o}_t, \mathbf{o}_{t+1}$. Both images are passed through the same backbone and neck to obtain $\mathbf{q}_t=g(f(\mathbf{o}_t))$ and $\mathbf{q}_{t+1}=g(f(\mathbf{o}_{t+1}))$, which are concatenated to make an action prediction via $\mathbf{y}_t = h(\mathbf{q}_t, \mathbf{q}_{t+1})$. Same as BC, we use the cross-entropy loss between $\mathbf{y}_t$ and $\mathbf{a}_t$.

\begin{table}[h]
\vspace{-3mm}
\begin{center}
\caption{Hyperparameters for pre-training IDM.}
\vspace{2mm}
\small{
\begin{tabular}{lr}
\toprule
Hyperparameter & Value \\
\midrule \\[-2.5ex]
Epochs (early stop)      & 100 (30) \\
Base learning rate       & 3e-4 \\
Weight decay             & 1e-5 \\
Optimizer ($\beta_1$, $\beta_2$)
                         & AdamW (0.9, 0.999) \\
Batch size               & 512 \\
\bottomrule
\end{tabular}}
\label{table:idm_hyperparameter}
\end{center}
\vspace{-3mm}
\end{table}

\subsubsection{SPR+IDM}
In combining the two algorithms, we do not make any modifications, and simply use the sum of two losses to pre-train the model.

\begin{table}[!h]
\vspace{-3mm}
\begin{center}
\caption{Hyperparameters for pre-training SPR+IDM.}
\vspace{2mm}
\small{
\begin{tabular}{lr}
\toprule
Hyperparameter & Value \\
\midrule \\[-2.5ex]
Epochs                   & 25 \\
Base learning rate       & 3e-5 \\
Weight decay             & 1e-5 \\
Optimizer ($\beta_1$, $\beta_2$)
                         & AdamW (0.9, 0.999) \\
Batch size               & 128 \\
\midrule \\[-2.5ex]
Prediction sequence length ($K$) & 4 \\
Momentum update ratio ($\tau$)
                         & [0.9, 0.999] \\
Representation dimensions
                         & 512 \\
Similarity measure       & Dot product \\
\bottomrule
\end{tabular}}
\label{table:spridm_hyperparameter}
\end{center}
\vspace{-3mm}
\end{table}



\newpage

\subsubsection{CQL}
Conservative Q-Learning \citep{kumar2020cql} is developed for learning Q values in an offline setting. We implement two distinct Q learning methods: Mean Squared Error (MSE) learning and Cross-entropy (Distributional) learning, both based on CQL.  We choose ResNet50 as the backbone architecture, with a neck structure consistent with that in Behavioral Cloning (BC) for both MSE and Distributional.

\textbf{MSE:} In MSE-based Q-learning, we employ a game-wise head that yields an output in the shape of $\mathbb{R}^{B \times T \times A}$, where $B$ denotes the batch size, $T$ represents the number of time steps, and $A$ signifies the size of the action space. We balance the MSE loss and CQL loss using a coefficient of 0.1.
\begin{table}[h]
\vspace{-3mm}
\begin{center}
\caption{Hyperparameters for pre-training CQL-M.}
\vspace{2mm}
\small{
\begin{tabular}{lr}
\toprule
Hyperparameter & Value \\
\midrule \\[-2.5ex]
Epochs                   & 100 \\
Base learning rate       & 1e-4 \\
Weight decay             & 1e-5 \\
Optimizer ($\beta_1$, $\beta_2$)
                         & AdamW (0.9, 0.95) \\
Batch size               & 512 \\
\midrule \\[-2.5ex]
CQL coefficient  & 0.1 \\
\bottomrule
\end{tabular}}
\label{table:cqlm_hyperparameter}
\end{center}
\vspace{-3mm}
\end{table}

\textbf{Distributional:} For cross-entropy-based distributional Q-learning, we utilize a game-wise head designed to produce an output dimension of $\mathbb{R}^{B \times T \times A \times N_A}$, with $B, T, A$ having the same implications as in MSE, and $N_A$ representing the number of atoms. We set a coefficient of 0.1 to balance cross-entropy loss and CQL loss, consistent with \citet{kumar2022sql}. Additionally, we adopted a support set of [-10, 10] following the methodology in \citet{bellemare2017distributional}.

\begin{table}[h]
\vspace{-3mm}
\begin{center}
\caption{Hyperparameters for pre-training CQL-D.}
\vspace{2mm}
\small{
\begin{tabular}{lr}
\toprule
Hyperparameter & Value \\
\midrule \\[-2.5ex]
Epochs                   & 100 \\
Base learning rate       & 1e-4 \\
Weight decay             & 1e-5 \\
Optimizer ($\beta_1$, $\beta_2$)
                         & AdamW (0.9, 0.95) \\
Batch size               & 512 \\
\midrule \\[-2.5ex]
Support Set & [-10, 10] \\
CQL coefficient  & 0.1 \\
\bottomrule
\end{tabular}}
\label{table:cqld_hyperparameter}
\end{center}
\vspace{-3mm}
\end{table}


\subsubsection{DT}

Decision Transformer (DT) \cite{chen2021dectransformer} adopts a unique perspective by treating reinforcement learning as a sequence modeling problem. It represents each transition within a trajectory as a triplet consisting of total return $\hat{R}_{t}$, observation $o_t$, and action $a_t$.  DT is trained to autoregressively predict these sequences, given the transition history $\bigcup^{t-1}_{i=1}\{\hat{R}_i, o_i, a_i\}$. At inference, this knowledge is leveraged to predict the optimal actions necessary for achieving a desired cumulative reward.

In our implementation, observations are encoded by the backbone and a 2-layer MLP with game-specific spatial embedding, generating $512$-dimensional tokens. Actions and returns are separately embedded into $512$-dimensional tokens, resulting in a sequence of length $K\times 3$. This sequence is processed through a causal transformer to produce $K$ outputs. The outputs are then utilized by the head layer to predict actions $a_{1:K}$. We employ cross-entropy loss for training the model.

\begin{table}[h]
\vspace{-3mm}
\begin{center}
\caption{Hyperparameters for pre-training DT.}
\vspace{2mm}
\small{
\begin{tabular}{lr}
\toprule
Hyperparameter & Value \\
\midrule \\[-2.5ex]
Epochs                   & 12 \\
Base learning rate       & 1e-4 \\
Weight decay             & 5e-2 \\
Optimizer ($\beta_1$, $\beta_2$)
                         & AdamW (0.9, 0.95) \\
Batch size               & 64 \\
\midrule \\[-2.5ex]
Sequence length (in tokens)    & $8\times 3$ \\
Reward scale             & 0.01 \\ 
Transformer embedding dimensions & 512 \\
Transformer MLP ratio    & 4 \\
Transformer heads        & 4 \\
Transformer layers       & 4 \\
\bottomrule
\end{tabular}}
\label{table:dt_hyperparameter}
\end{center}
\vspace{-3mm}
\end{table}


\newpage

\subsection{Fine-tuning}
After pre-training, we assess the pre-trained models on two downstream adaptation scenarios: Offline Behavioral Cloning (BC) and Online Reinforcement Learning (RL). Each task is conducted on three distinct sets of games (ID, Near-OOD, Far-OOD), including two sets unseen during pre-training. Results are compiled from three independent runs (seeds).

The backbone network is kept frozen to focus on the quality of representations, while other components are re-initialized. Any modifications revert to the standard architecture as detailed in Section \ref{sub_section:model} and Section \ref{appendix:impl_pretrain}. This process is uniformly applied across all games.

\subsubsection{Offline BC}
In Offline BC, expert demonstrations are used for learning control. Specifically, the fine-tuning dataset for ID and Near-OOD games are sampled from DQN-Replay-Dataset. From the five runs, we use the last checkpoints and sample the initial 10,000 interactions. For Far-OOD games, we train a Rainbow agent on each environment for 2M steps and record the last 50,000 interactions. As a result, each fine-tuning stage is performed over a 50k dataset for 100 epochs. The final performance is evaluated by the average gameplay score of 100 trials.

\begin{table}[h]
\vspace{-3mm}
\begin{center}
\caption{Hyperparameters for downstream Offline BC.}
\vspace{2mm}
\small{
\begin{tabular}{lr}
\toprule
Hyperparameter & Value \\
\midrule \\[-2.5ex]
Augmentation             & [Random Shift, Intensity] \\
Random shift pad         & 4 \\
Intensity scale          & 0.05 \\
\midrule \\[-2.5ex]
Learning rate scheduler  & Cosine annealing with warmup \\
Warmup ratio             & 0.1 \\
Initial learning rate ratio & 0.1 \\
Base learning rate       & 1e-3 \\
Weight decay             & 1e-4 \\
Optimizer ($\beta_1$, $\beta_2$)
                         & AdamW (0.9, 0.999) \\
Batch size               & 512 \\
\bottomrule
\end{tabular}}
\label{table:offlinebc_hyperparameter}
\end{center}
\vspace{-3mm}
\end{table}

\newpage

\subsubsection{Online RL}

Following the standard experimental setup from sample-efficient Atari benchmark \cite{kaiser2019model, schwarzer2021sgi}, we train a Q-learning layer on top of the frozen encoder via the Rainbow algorithm \cite{hessel2018rainbow}.
For the sake of simplicity, we do not use noisy layers and use epsilon greedy for exploration.
Similar to the Offline BC, we trained the model for 50k steps and the performance is measured by the average gameplay score over 100 attempts.
Detailed hyperparameters are listed in Table \ref{table:onlinerl_hyperparameter}. 

\begin{table}[h]
\vspace{-3mm}
\begin{center}
\caption{Hyperparameters for downstream Online RL.}
\vspace{2mm}
\small{
\begin{tabular}{lr}
\toprule
Hyperparameter & Value \\
\midrule \\[-2.5ex]
Augmentation             & [Random Shift, Intensity] \\
Random shift pad         & 4 \\
Intensity scale          & 0.05 \\
\midrule \\[-2.5ex]

Training steps           & 50k \\
Update                   & Distributional Q \\
Dueling                  & True \\
Support of Q-distribution& 51 \\
Discount factor $\gamma$ & 0.99 \\
Batch size               & 32  \\
Optimizer ($\beta_1$, $\beta_2$, $\epsilon$) 
                         & Adam (0.9, 0.999, 0.000015) \\
Learning rate            & 0.0001 \\
Max gradient norm        & 10 \\
Priority exponent        & 0.5 \\
Priority correction      & 0.4 $\rightarrow$ 1 \\
Exploration Schedule (start, end, steps)
                         & Epsilon Greedy (50k, 1.0, 0.02) \\
Replay buffer size       & 50k \\
Min buffer size for sampling & 2000 \\
Replay per training step & 1 \\
Updates per replay step  & 2 \\
Multi-step return length & 10 \\
Q-head hidden units      & 1024 \\
Q-head non-linearity     & ReLU \\
Evaluation trajectories  & 100 \\
\bottomrule
\end{tabular}}
\label{table:onlinerl_hyperparameter}
\end{center}
\vspace{-3mm}
\end{table}


\clearpage


\section{Extended Experimental Results}
\label{supp:extended_mian_result}
Here, we present additional experimental results that were not included in the Section \ref{sec:main_result}.

\subsection{Game-Mode Change}
\label{supp:gamemode}
For Near-OOD in the main experiment (Figure \ref{figure:main}), we fine-tuned with offline BC and online RL using games that were different from the pre-training but belonged to the same task genre. However, some might think that simply changing the modes of the games used in pre-training would be more suitable for Near-OOD \cite{machado2018atari}. The reason we chose the Near-OOD environment in the main experiment is that changing the mode results in slight changes, such as alterations in the color or shape of objects. Therefore, we thought it was unrealistic because it is very similar to the pre-training environment, indicating that the fine-tuning environment is almost the same as the pre-training environment. For example, let's assume we are training a house-cleaning robot using the pretrain-then-finetune method. In this case, changing the mode is analogous to training the robot on different colored/shaped objects within the same house in the pre-training stage. Assuming that the fine-tuning environment is the same as the pre-training environment is unrealistic.

\begin{figure}
     \centering
     \subfloat[Aggregate scores]{
        \includegraphics[width=0.31\linewidth]{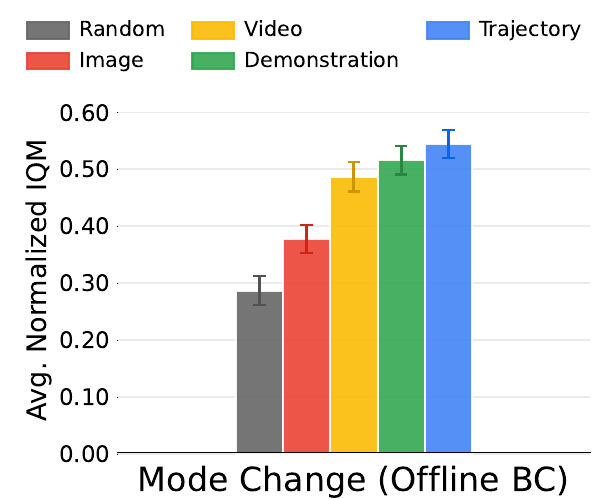}
    }
     \subfloat[Algorithm-wise Scores]{
        \includegraphics[width=0.69\linewidth]{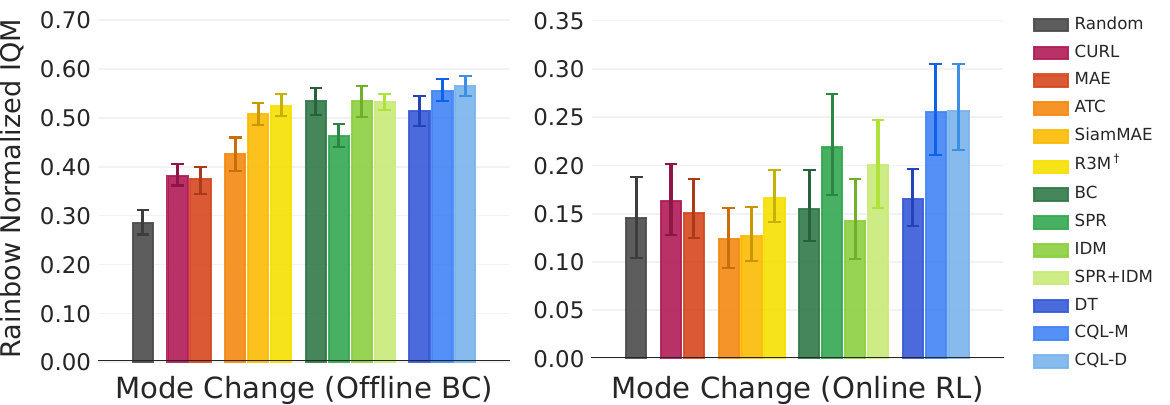}
    }
    \caption{The results of offline BC and online RL in environments where only the mode is changed from the pre-training game environments.}
    \label{fig:change_game_result}
\end{figure}

To support our selection, we conducted offline BC and online RL fine-tuning experiments by changing the modes of the games listed in Table \ref{table:mode_change_game}. Additionally, to match the scale of the results for each game, we normalized with the final scores of the Rainbow agent trained with 2 million steps as in the Far-OOD setting of the main experiment. The experimental results are shown in Figure \ref{fig:change_game_result}. Similar to Figure \ref{figure:main}(a), learning reward-specific knowledge yields best performance even in the Game-Mode Change environment. This means that the reward function in the Game-Mode Change environment is very similar to the pre-training environment. In other words, simply changing the game mode is not significantly different from the ID.

\begin{table}[h]
\vspace{-3mm}
\begin{center}
\caption{Games used in the game mode change experiment. }
\label{table:mode_change_game}
\vspace{0.8ex}
\small
{
\begin{tabular}{l| p{10cm}}
\toprule

Games(Mode)  & AirRaid(5), Asteroids(17), Atlantis(3), BankHeist(5), Berzerk(7), \newline
        Breakout(7), CrazyClimber(3), DemonAttack(3), DoubleDunk(9), Freeway(5), \newline
        Gravitar(3), Hero(3), Krull(3), MsPacman(3), PrivateEye(3), \newline
        Skiing(6), SpaceInvaders(9), StarGunner(3), Tutankham(3), YarsRevenge(3) \newline
        Zaxxon(3) \\
\bottomrule
\end{tabular}}
\end{center}
\end{table}

\subsection{Relationship Between Object Size and Mask-Based Methods}
\label{supp:obj_size}


One unique property of the Atari environments is the large variance of object sizes, ranging from few hundred pixels to mere 1-2 pixels. This led us to hypothesize that mask-based reconstruction methods may struggle to deal with small objects, as these objects can be entirely occluded by masking operation. 

\begin{figure}[h]
\begin{center}
\includegraphics[width=0.60\linewidth]{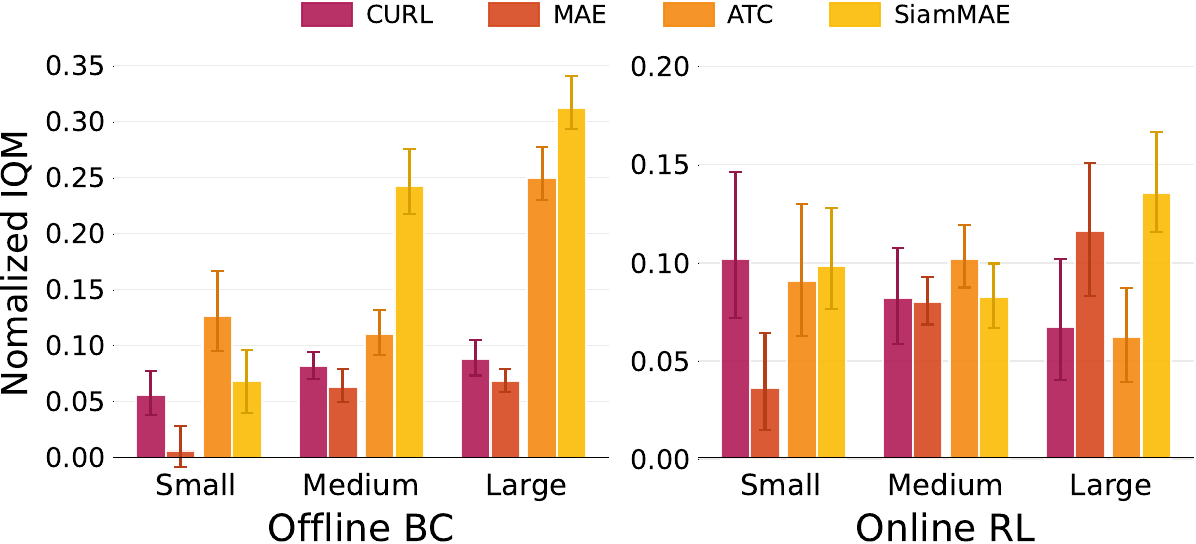}
\end{center}
\vspace{-2mm}
\caption{\textbf{Mask-based methods relative to object sizes.} A comparative analysis of mask-based versus non-mask methods in ID and Near-OOD games, classified by object size (small, medium, large). The results indicate that mask-based methods tend to excel more in games with larger objects compared to smaller ones. }
\label{figure:obj_size}
\end{figure}

To investigate this, we categorized ID and Near-OOD environments by their object size (see Table \ref{table:game_object_size}) and compared the mask-based methods (MAE, SiamMAE) against latent reconstruction methods in the same category (CURL, ATC).
As illustrated in Figure \ref{figure:obj_size}, the results indeed support our proposition.
Mask-based methods underperformed with small objects and excelled with larger objects, implying that the object size in environments can play a crucial role in these models.

\begin{table}[h]
\vspace{-3mm}
\begin{center}
\caption{Games categorized by object size. Far-OOD games are not included. }
\label{table:game_object_size}
\vspace{0.8ex}
\small
{
\begin{tabular}{l p{10cm}}
\toprule
Object size & Games \\
\midrule \\[-2.7ex]
Small  & Amidar, BankHeist, Bowling, Breakout, Centipede, \newline
         Gravitar, Jamesbond, Krull, MsPacman, Phoenix, \newline
         Pitfall, Pong, Pooyan, Riverraid, Skiing, \newline
         StarGunner, TimePilot, Tutankham, Venture, VideoPinball  \\
\midrule
Medium & Alien, Asterix, Asteroids, Atlantis, BeamRider, \newline
         Berzerk, Carnival, ChopperCommand, CrazyClimber, DemonAttack, \newline
         DoubleDunk, ElevatorAction, FishingDerby, Freeway, Frostbite, \newline
         Hero, IceHockey, JourneyEscape, Kangaroo, MontezumaRevenge, \newline
         NameThisGame, PrivateEye, Qbert, Robotank, Seaquest \newline
         Solaris, SpaceInvaders, Tennis, WizardOfWor, YarsRevenge \\
\midrule
Large  & AirRaid, Assault, BattleZone, Boxing, Enduro, \newline
         Gopher, KungFuMaster, RoadRunner, UpNDown, Zaxxon\\
\bottomrule
\end{tabular}}
\end{center}
\end{table}

\newpage

\subsection{Eigen-CAM Analysis} 
\label{supp:eigen_cam}

In this section, we analyze the pre-trained backbones based on what aspects of the observations they concentrate on.
We employed Eigen-CAM \citep{muhammad2020eigencam}, a visualization method frequently used in computer vision and deep learning.
We randomly sampled an observation from two In-Distribution and two Near-Out-of-Distribution games, and applied Eigen-CAM on our main models in Section \ref{sec:main_result}. 

\subsubsection{In-Distribution environment}
\begin{figure}[H]
    \centering
    \subfloat{
        \includegraphics[width=0.9\textwidth]{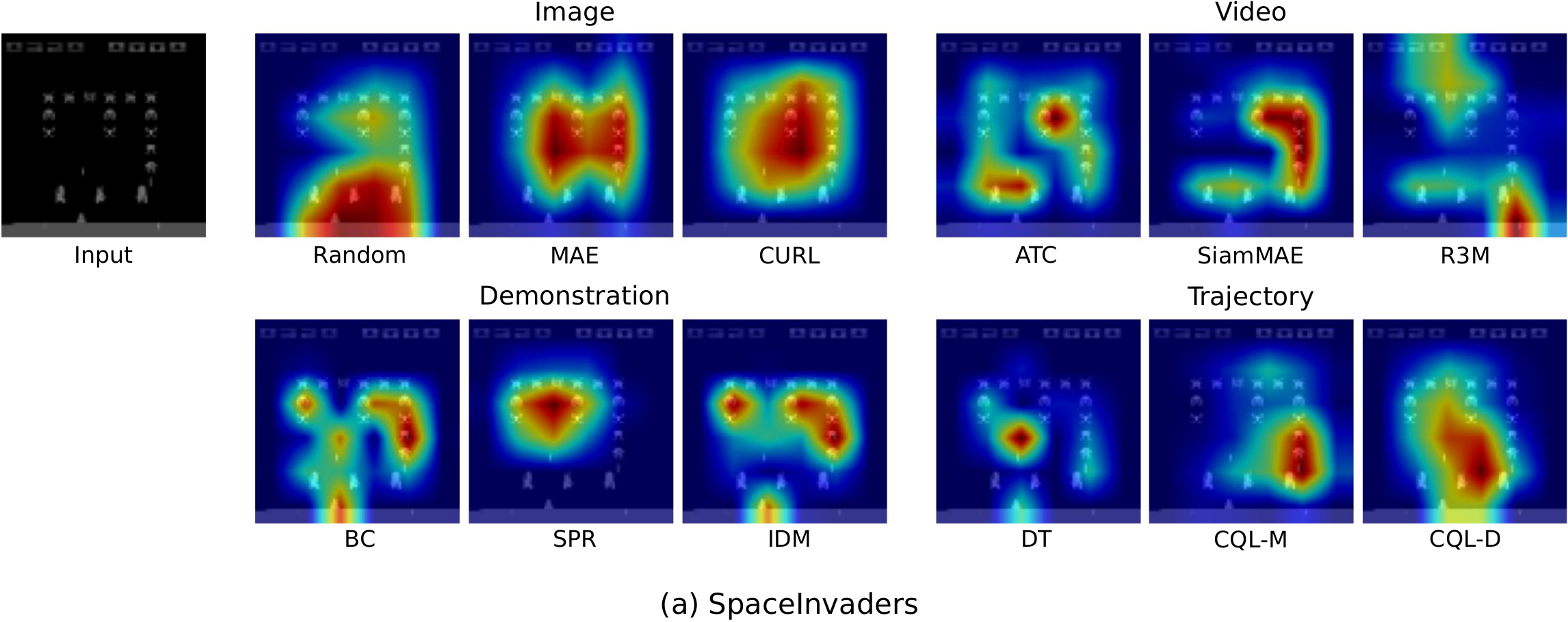}
    }
    \\
    \subfloat{
        \includegraphics[width=0.9\textwidth]{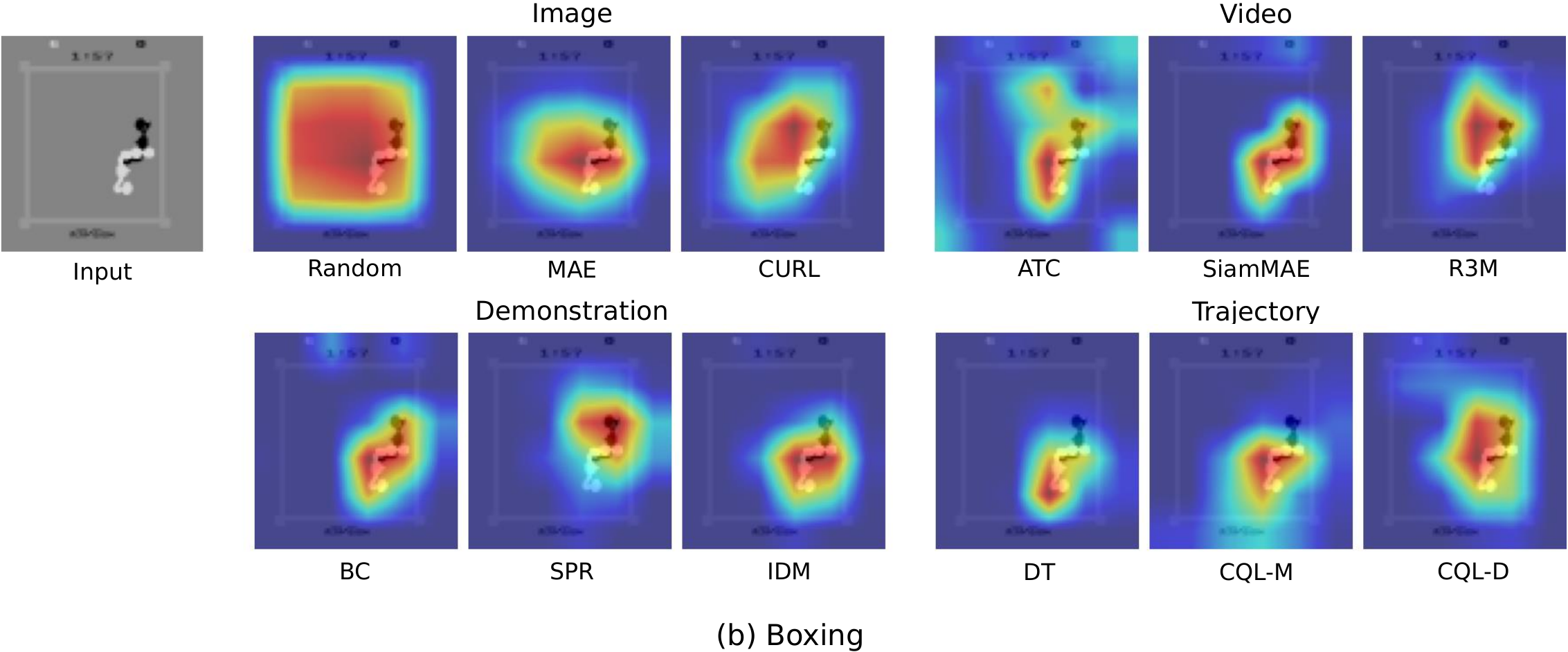}
    }
    \caption{\textbf{Eigen-CAM analysis of ID games.} (a) Space Invaders is a game in which players are required to evade enemy attacks while eliminating as many adversaries as possible through vertical shooting. (b) Boxing is a game where the player aims to score more points than their opponent by striking the opponent's face as many as possible within a set time limit. When looking at Eigen-CAM images, it's evident that encoders pre-trained with temporal dynamics and task-related information such as video, action, and reward, rather than just spatial information from images, are more effective in identifying objects relevant to the task. Additionally, encoders trained using demonstration data, which includes task-relevant information, show improved ability in capturing and understanding important features for the task.
    }
    \label{figure:grad_cam_id}
\end{figure}

For In-Distribution environments, we employ 'Space Invaders' and 'Boxing'. 
The outcomes of this analysis are illustrated in Figure \ref{figure:grad_cam_id}. 
It was observed that when the encoder is pre-trained solely with images, it struggles to capture objects relevant to the task. 
However, when pre-trained with data enriched with additional information such as temporal dynamics and task-relevant information, and values of states and state-action pairs, the encoder effectively identifies meaningful objects.

\subsubsection{Near-Out-of-Distribution environment}
\begin{figure}[H]
    \centering
    \subfloat{
        \includegraphics[width=0.9\textwidth]{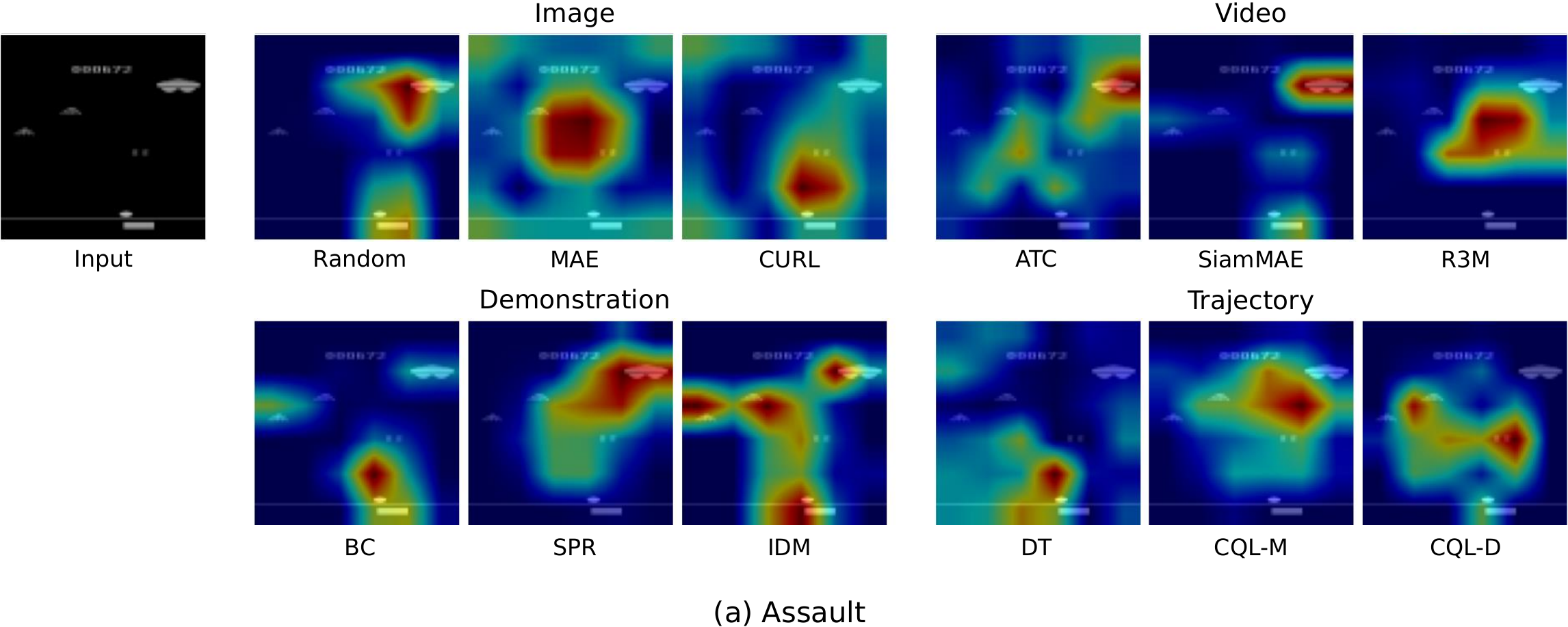}
    }
    \\
    \subfloat{
        \includegraphics[width=0.9\textwidth]{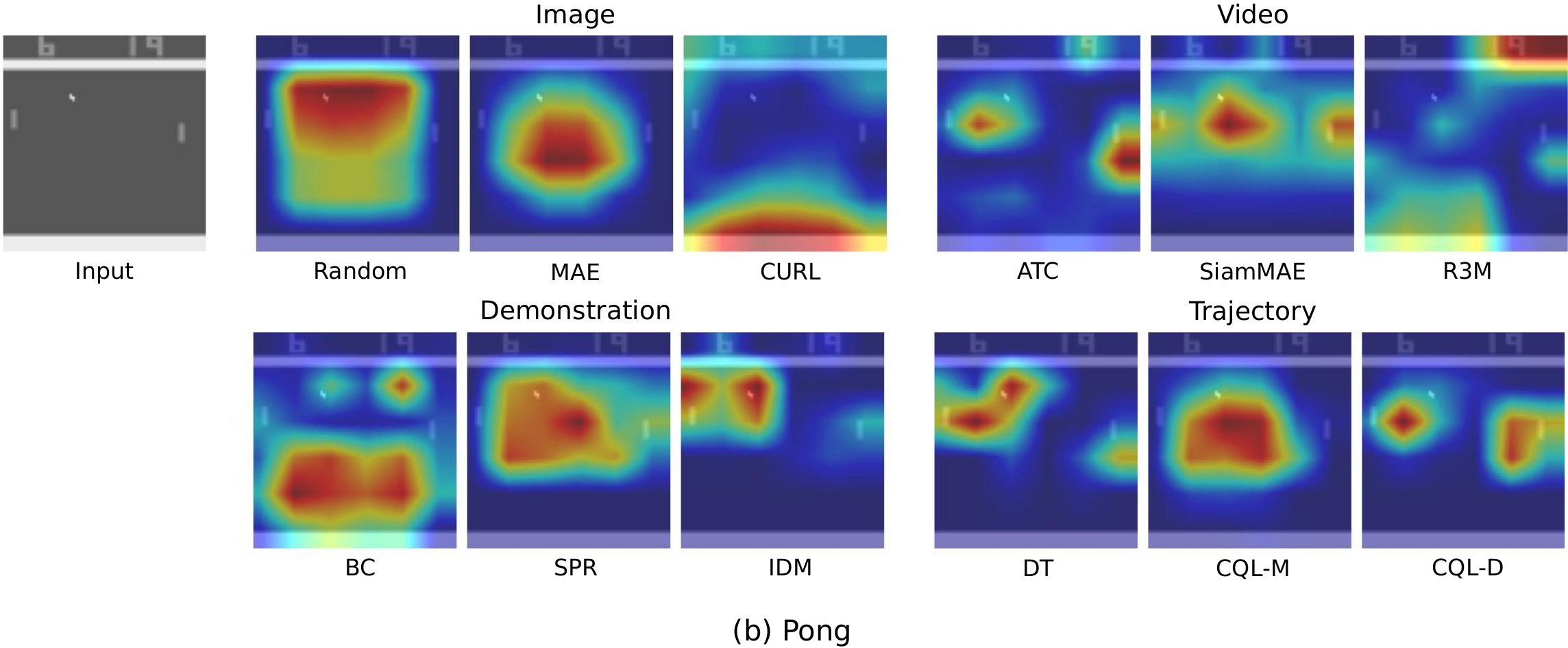}
    }
    \caption{\textbf{Eigen-CAM analysis of Near-OOD games.} (a) Assault is a game where players must avoid enemy assaults and eliminate as many opponents as possible through either vertical or horizontal shooting. (b) Pong is a game where the player who reaches a specified score wins by scoring points while preventing the opponent from hitting the ball. Similar to what we observe in ID, encoders that are trained on temporal dynamics, task-related information, and the value between states and actions using videos, demonstrations, and trajectories demonstrate superior ability in recognizing task-related objects, compared to those that only learn spatial aspects from images.}
    \label{figure:grad_cam_ood}
\end{figure}

We also analyze the pre-trained encoder in an out-of-distribution environment, which was not used during its pre-training phase.
Specifically, 'Assault' and 'Pong' are utilized for this purpose. 
The outcomes are shown in Figure \ref{figure:grad_cam_ood}. 
Similar to the situation with an in-distribution environment, encoders pre-trained on images alone had difficulty in targeting significant objects, while those pre-trained on a richer dataset demonstrated effective identification of important objects.

\section{Full Results}
We provide the individual game scores of all algorithms in our experiments. The scores are recorded at the end of fine-tuning and averaged over 3 seeds. We also include the mean scores of a randomly behaving agent (aliased \textit{RndmAgnt}) and the DQN scores reported by \citet{castro18dopamine}, which are used to compute the DQN normlized score with the formula: $s_{\text{DNS}}=(s_{\text{agent}}-s_{\text{min}})/(s_{\text{max}}-s_{\text{min}})$ where $s_{\text{max}} = \max(s_{\textit{DQN}},s_{\textit{RndmAgnt}}), s_{\text{min}} = \min(s_{\textit{DQN}},s_{\textit{RndmAgnt}})$, and $s_{\text{agent}}$ is the score to be normalized. For Far-OOD games, instead of DQN, we report the scores of Rainbow \citep{hessel2018rainbow} trained for 2M steps and use them for normalization. In addition to IQM, we report the optimality gap of the normalized scores.

\subsection{Main Results}

\begin{table}[!h]
\vspace{-2.5mm}
\begin{center}
\caption{Mean episodic scores of Offline BC on In-Distribution games.}
\vspace{-2.0ex}
\resizebox{0.99\textwidth}{!}{
\begin{tabular}{r|cc|c|cc|ccc|ccc|ccc}
\toprule &\\[-2.5ex]
Game & \textit{RndmAgnt} & \textit{DQN} & Random & MAE & CURL & ATC & SiamMAE & R3M$^\dagger$ & BC & SPR & IDM & DT & CQL-M & CQL-D \\[0.2ex]
\midrule \\[-2.7ex]
AirRaid           & $    579.2$ & $   7479.5$ & $   1262.2$ & $    973.8$ & $    970.1$ & $   1479.6$ & $   3106.9$ & $   1422.1$ & $   4290.6$ & $   1521.2$ & $   2743.8$ & $   3497.9$ & $   3189.5$ & $   3754.3$ \\
Amidar            & $      5.8$ & $   1207.7$ & $     55.5$ & $     60.1$ & $     28.9$ & $     60.3$ & $     34.8$ & $     44.8$ & $     65.0$ & $     58.0$ & $     56.8$ & $     58.6$ & $     69.4$ & $     78.5$ \\
Asteroids         & $    719.1$ & $    698.4$ & $    668.2$ & $    603.3$ & $    635.1$ & $    693.6$ & $    826.6$ & $    800.4$ & $    671.1$ & $    872.8$ & $    724.8$ & $    812.2$ & $    794.3$ & $    998.3$ \\
Atlantis          & $  12850.0$ & $ 853640.0$ & $   7607.7$ & $  18895.3$ & $  12785.7$ & $  13607.0$ & $  49505.7$ & $  29909.7$ & $  24181.3$ & $ 472835.3$ & $  48420.7$ & $  25266.7$ & $  22112.7$ & $  16246.3$ \\
BankHeist         & $     14.2$ & $    601.8$ & $      0.0$ & $     18.2$ & $     32.9$ & $      8.5$ & $     21.7$ & $     23.8$ & $     23.4$ & $     40.9$ & $     26.1$ & $     30.6$ & $     62.3$ & $     70.8$ \\
\bottomrule
\end{tabular}}
\end{center}
\label{table:full_result_offlinebc_id}
\end{table}

\begin{table}[!h]
\vspace{-3.5mm}
\begin{center}
\caption{Mean episodic scores of Offline BC on Near-Out-of-Distribution games.}
\vspace{-2.0ex}
\resizebox{0.99\textwidth}{!}{
\begin{tabular}{r|cc|c|cc|ccc|ccc|ccc}
\toprule &\\[-2.5ex]
Game & \textit{RndmAgnt} & \textit{DQN} & Random & MAE & CURL & ATC & SiamMAE & R3M$^\dagger$ & BC & SPR & IDM & DT & CQL-M & CQL-D \\[0.2ex]
\midrule \\[-2.7ex]
Alien          & $    227.8$ & $   2484.5$ & $    553.5$ & $    381.3$ & $    536.6$ & $    590.8$ & $    863.0$ & $    539.0$ & $    727.0$ & $    701.1$ & $    816.8$ & $    797.8$ & $    831.6$ & $    469.8$ \\
Assault        & $    222.4$ & $   1525.2$ & $    414.6$ & $    134.0$ & $    396.8$ & $    623.0$ & $    726.7$ & $    618.4$ & $    758.5$ & $    550.0$ & $    751.1$ & $    795.0$ & $    738.2$ & $    702.7$ \\
Asterix        & $    210.0$ & $   2711.4$ & $    281.5$ & $    236.8$ & $    294.8$ & $    365.3$ & $    349.5$ & $    392.8$ & $    432.7$ & $    347.5$ & $    487.7$ & $    470.0$ & $    421.7$ & $    521.2$ \\
BeamRider      & $    363.9$ & $   5852.4$ & $    548.3$ & $    522.7$ & $    533.9$ & $    543.5$ & $    523.6$ & $    623.4$ & $    554.3$ & $    471.0$ & $    551.1$ & $    574.4$ & $    535.9$ & $    519.5$ \\
JourneyEscape  & $ -19977.0$ & $  -3671.1$ & $  -8998.0$ & $ -12759.3$ & $ -14779.0$ & $ -11430.0$ & $ -11594.0$ & $ -12088.7$ & $ -11677.3$ & $ -14696.3$ & $ -13380.0$ & $ -11251.3$ & $ -12415.3$ & $ -12341.3$ \\
\bottomrule
\end{tabular}}
\end{center}
\label{table:full_result_offlinebc_near_ood}
\end{table}

\begin{table}[!h]
\vspace{-3.5mm}
\begin{center}
\caption{Mean episodic scores of Offline BC on Far-Out-of-Distribution games.}
\vspace{-2.0ex}
\resizebox{0.99\textwidth}{!}{
\begin{tabular}{r|cc|c|cc|ccc|ccc|ccc}
\toprule &\\[-2.5ex]
Game & \textit{RndmAgnt} & \textit{Rainbow} & Random & MAE & CURL & ATC & SiamMAE & R3M$^\dagger$ & BC & SPR & IDM & DT & CQL-M & CQL-D \\[0.2ex]
\midrule \\[-2.7ex]
BasicMath        & $   0.1334$ & $   3.5700$ & $   0.6300$ & $   1.3700$ & $   2.4500$ & $   2.8633$ & $   3.4967$ & $   3.4000$ & $   2.7200$ & $   2.9467$ & $   3.5867$ & $   2.5167$ & $   2.8167$ & $   3.3300$ \\
HumanCannonball  & $   1.0767$ & $   5.5600$ & $   0.7067$ & $   1.5033$ & $   1.1533$ & $   2.8133$ & $   1.0500$ & $   1.0267$ & $   1.8133$ & $   0.8167$ & $   1.1900$ & $   1.0600$ & $   1.1833$ & $   1.3433$ \\
Klax             & $   0.3734$ & $2124.7500$ & $   4.4000$ & $   3.4167$ & $   6.7667$ & $  12.8333$ & $  13.4333$ & $  66.8167$ & $  40.6667$ & $  23.1000$ & $  25.2667$ & $  47.8500$ & $  28.2833$ & $  21.4667$ \\
Othello          & $ -21.3233$ & $  -2.0100$ & $  -0.0067$ & $   0.0167$ & $  -0.0200$ & $  -0.1500$ & $   0.1167$ & $   0.0767$ & $   0.0067$ & $   0.0667$ & $   0.0867$ & $   0.0933$ & $   0.1933$ & $   0.1333$ \\
Surround         & $  -9.9833$ & $  -7.8100$ & $  -9.9967$ & $  -9.2533$ & $  -9.2800$ & $  -9.0533$ & $  -8.9067$ & $  -8.9833$ & $  -9.1100$ & $  -9.2300$ & $  -8.7600$ & $  -8.9567$ & $  -9.2400$ & $  -9.0233$ \\
\bottomrule
\end{tabular}}
\end{center}
\label{table:full_result_offlinebc_far_ood}
\end{table}

\newpage

\begin{table}[!h]
\vspace{0mm}
\begin{center}
\caption{Mean episodic scores of Online RL on In-Distribution games.}
\vspace{-2.0ex}
\resizebox{0.99\textwidth}{!}{
\begin{tabular}{r|cc|c|cc|ccc|ccc|ccc}
\toprule &\\[-2.5ex]
Game & \textit{RndmAgnt} & \textit{DQN} & Random & MAE & CURL & ATC & SiamMAE & R3M$^\dagger$ & BC & SPR & IDM & DT & CQL-M & CQL-D \\[0.2ex]
\midrule \\[-2.7ex]
AirRaid           & $    579.2$ & $   7479.5$ & $    368.2$ & $    601.1$ & $    502.8$ & $    718.0$ & $    639.6$ & $    570.6$ & $    803.7$ & $    512.7$ & $    636.4$ & $    782.2$ & $   1087.8$ & $    848.9$ \\
Amidar            & $      5.8$ & $   1207.7$ & $    131.0$ & $    110.3$ & $    105.3$ & $    109.0$ & $    105.8$ & $     95.7$ & $    165.3$ & $     83.1$ & $    111.2$ & $    105.5$ & $    116.5$ & $    128.3$ \\
Asteroids         & $    719.1$ & $    698.4$ & $   1008.3$ & $    739.4$ & $    755.6$ & $    811.5$ & $    751.2$ & $    718.6$ & $    701.2$ & $    918.4$ & $    894.5$ & $    857.7$ & $    821.3$ & $   1061.6$ \\
Atlantis          & $  12850.0$ & $ 853640.0$ & $  10404.7$ & $  17912.0$ & $  29700.7$ & $  14236.7$ & $  18619.3$ & $  24911.0$ & $  19350.0$ & $  79626.0$ & $  18266.3$ & $  30214.7$ & $  31849.7$ & $  38407.3$ \\
BankHeist         & $     14.2$ & $    601.8$ & $      9.7$ & $     40.9$ & $     37.3$ & $     39.9$ & $     54.8$ & $     56.1$ & $    357.1$ & $     57.7$ & $     63.9$ & $     47.8$ & $    300.3$ & $    569.5$ \\
\bottomrule
\end{tabular}}
\end{center}
\label{table:full_result_onlinerl_id}
\end{table}

\begin{table}[!h]
\vspace{-3.5mm}
\begin{center}
\caption{Mean episodic scores of Online RL on Near-Out-of-Distribution games.}
\vspace{-2.0ex}
\resizebox{0.99\textwidth}{!}{
\begin{tabular}{r|cc|c|cc|ccc|ccc|ccc}
\toprule &\\[-2.5ex]
Game & \textit{RndmAgnt} & \textit{DQN} & Random & MAE & CURL & ATC & SiamMAE & R3M$^\dagger$ & BC & SPR & IDM & DT & CQL-M & CQL-D \\[0.2ex]
\midrule \\[-2.7ex]
Alien          & $    227.8$ & $   2484.5$ & $    444.2$ & $    641.9$ & $    595.0$ & $    659.7$ & $    566.0$ & $    667.2$ & $    772.9$ & $    604.3$ & $    546.3$ & $    538.0$ & $    640.8$ & $    621.0$ \\
Assault        & $    222.4$ & $   1525.2$ & $    607.7$ & $    572.8$ & $    439.8$ & $    453.9$ & $    564.7$ & $    379.3$ & $    371.4$ & $    510.3$ & $    334.9$ & $    351.3$ & $    545.9$ & $    461.1$ \\
Asterix        & $    210.0$ & $   2711.4$ & $    536.2$ & $    476.0$ & $    533.8$ & $    485.2$ & $    573.8$ & $    496.7$ & $    499.2$ & $    571.8$ & $    608.0$ & $    396.5$ & $    615.2$ & $    545.2$ \\
BeamRider      & $    363.9$ & $   5852.4$ & $    581.1$ & $    479.0$ & $    597.5$ & $    542.4$ & $    575.9$ & $    701.5$ & $    598.2$ & $    510.2$ & $    562.7$ & $    560.6$ & $    674.4$ & $    591.5$ \\
JourneyEscape  & $ -19977.0$ & $  -3671.1$ & $  -6439.0$ & $ -12185.3$ & $  -7828.3$ & $ -11409.0$ & $ -10866.3$ & $ -11459.3$ & $  -9623.0$ & $ -11564.3$ & $ -14326.0$ & $ -17966.0$ & $  -9327.0$ & $  -8893.0$ \\
\bottomrule
\end{tabular}}
\end{center}
\label{table:full_result_onlinerl_near_ood}
\end{table}

\begin{table}[!h]
\vspace{-3.5mm}
\begin{center}
\caption{Mean episodic scores of Online RL on Far-Out-of-Distribution games.}
\vspace{-2.0ex}
\resizebox{0.99\textwidth}{!}{
\begin{tabular}{r|cc|c|cc|ccc|ccc|ccc}
\toprule &\\[-2.5ex]
Game & \textit{RndmAgnt} & \textit{Rainbow} & Random & MAE & CURL & ATC & SiamMAE & R3M$^\dagger$ & BC & SPR & IDM & DT & CQL-M & CQL-D \\[0.2ex]
\midrule \\[-2.7ex]
BasicMath        & $   0.1334$ & $   3.5700$ & $   0.0000$ & $   0.2233$ & $   0.8700$ & $   0.0967$ & $   0.3367$ & $   0.2400$ & $   0.0900$ & $   0.2667$ & $   0.6533$ & $   0.9233$ & $   0.0700$ & $   0.0633$ \\
HumanCannonball  & $   1.0767$ & $   5.5600$ & $   1.0833$ & $   1.0500$ & $   1.4633$ & $   1.6667$ & $   2.0400$ & $   1.6767$ & $   1.5033$ & $   1.3400$ & $   1.9767$ & $   1.4967$ & $   1.6667$ & $   1.4400$ \\
Klax             & $   0.3734$ & $2124.7500$ & $   5.1667$ & $   6.0500$ & $  10.8000$ & $   7.5000$ & $  10.1000$ & $  13.2667$ & $  11.5333$ & $   9.3167$ & $   9.5667$ & $  10.4500$ & $  10.0833$ & $  10.1500$ \\
Othello          & $ -21.3233$ & $  -2.0100$ & $   0.0000$ & $   0.3400$ & $  -0.5067$ & $   0.0667$ & $  -0.0667$ & $  -0.1000$ & $  -0.6933$ & $  -0.1067$ & $   0.0200$ & $  -0.0400$ & $  -0.7400$ & $  -0.0667$ \\
Surround         & $  -9.9833$ & $  -7.8100$ & $  -9.9433$ & $  -9.8633$ & $  -9.7933$ & $  -9.8067$ & $  -9.8800$ & $  -9.6867$ & $  -9.7200$ & $  -9.7467$ & $  -9.7400$ & $  -9.8133$ & $  -9.8167$ & $  -9.8333$ \\
\bottomrule
\end{tabular}}
\end{center}
\label{table:full_result_onlinerl_far_ood}
\end{table}

\newpage


\subsection{Ablation Results}

\begin{table}[h]
\vspace{-2.5mm}
\begin{center}
\caption{Offline BC scores on In-Distribution games, after pre-training with datasets of differing optimality.} 
\vspace{-2.0ex}
\resizebox{0.75\textwidth}{!}{
\begin{tabular}{r|cc|cccc|cccc}
\multicolumn{3}{c}{} & \multicolumn{4}{c}{\textbf{Suboptimal}} &  \multicolumn{4}{c}{\textbf{Expert}} \\
\toprule &\\[-2.5ex]
Game & \textit{RndmAgnt} & \textit{DQN} & CURL & ATC & BC & CQL-D & CURL & ATC & BC & CQL-D \\[0.2ex]
\midrule \\[-2.7ex]
AirRaid           & $    579.2$ & $   7479.5$ & $   1216.8$ & $   1699.8$ & $   2955.7$ & $   3120.6$ & $   1213.8$ & $   1393.0$ & $   4803.5$ & $   3953.7$ \\
Amidar            & $      5.8$ & $   1207.7$ & $     44.1$ & $     37.5$ & $     68.6$ & $     63.2$ & $     47.7$ & $     44.6$ & $     67.1$ & $     92.5$ \\
Asteroids         & $    719.1$ & $    698.4$ & $    767.5$ & $    725.5$ & $    614.5$ & $    833.9$ & $    735.8$ & $    742.2$ & $    593.4$ & $    849.4$ \\
Atlantis          & $  12850.0$ & $ 853640.0$ & $  24057.3$ & $  24262.3$ & $  16286.3$ & $  15749.3$ & $  15637.3$ & $  18773.7$ & $  22173.0$ & $   9151.0$ \\
\bottomrule
 \end{tabular}}
\end{center}
\label{table:full_result_suboptimality_id}
\end{table}

\begin{table}[!h]
\vspace{-4.5mm}
\begin{center}
\caption{Offline BC scores on Near-Out-of-Distribution games, after pre-training with datasets of differing optimality.} 
\vspace{-2.0ex}
\resizebox{0.75\textwidth}{!}{
\begin{tabular}{r|cc|cccc|cccc}
\multicolumn{3}{c}{} & \multicolumn{4}{c}{\textbf{Suboptimal}} &  \multicolumn{4}{c}{\textbf{Expert}} \\
\toprule &\\[-2.5ex]
Game & \textit{RndmAgnt} & \textit{DQN} & CURL & ATC & BC & CQL-D & CURL & ATC & BC & CQL-D \\[0.2ex]
\midrule \\[-2.7ex]
Alien          & $    227.8$ & $   2484.5$ & $    659.6$ & $    469.6$ & $    902.8$ & $    466.6$ & $    574.4$ & $    588.8$ & $    750.2$ & $    693.2$ \\
Assault        & $    222.4$ & $   1525.2$ & $    485.7$ & $    698.3$ & $    680.1$ & $    578.0$ & $    501.4$ & $    656.0$ & $    801.7$ & $    680.7$ \\
Asterix        & $    210.0$ & $   2711.4$ & $    409.2$ & $    346.7$ & $    510.8$ & $    542.5$ & $    384.2$ & $    404.5$ & $    487.0$ & $    455.7$ \\
BeamRider      & $    363.9$ & $   5852.4$ & $    532.3$ & $    542.0$ & $    564.5$ & $    485.9$ & $    536.7$ & $    537.7$ & $    613.4$ & $    520.7$ \\
\bottomrule
\end{tabular}}
\end{center}
\label{table:full_result_suboptimality_near_ood}
\end{table}

\begin{table}[!h]
\vspace{-4.5mm}
\begin{center}
\caption{Offline BC scores on Far-Out-of-Distribution games, after pre-training with datasets of differing optimality.} 
\vspace{-2.0ex}
\resizebox{0.72\textwidth}{!}{
\begin{tabular}{r|cc|cccc|cccc}
\multicolumn{3}{c}{} & \multicolumn{4}{c}{\textbf{Suboptimal}} &  \multicolumn{4}{c}{\textbf{Expert}} \\
\toprule &\\[-2.5ex]
Game & \textit{RndmAgnt} & \textit{Rainbow} & CURL & ATC & BC & CQL-D & CURL & ATC & BC & CQL-D \\[0.2ex]
\midrule \\[-2.7ex]
BasicMath        & $   0.1334$ & $   3.5700$ & $   2.0433$ & $   3.3667$ & $   2.3867$ & $   2.8467$ & $   2.3200$ & $   3.1400$ & $   2.2800$ & $   2.3767$ \\
HumanCannonball  & $   1.0767$ & $   5.5600$ & $   1.6767$ & $   1.5600$ & $   1.4800$ & $   1.3667$ & $   1.0800$ & $   1.4067$ & $   1.4100$ & $   1.1767$ \\
Klax             & $   0.3734$ & $2124.7500$ & $  16.0167$ & $  19.5167$ & $  73.9000$ & $  23.6333$ & $  13.7000$ & $  20.7000$ & $  24.4000$ & $  20.0667$ \\
Othello          & $ -21.3233$ & $  -2.0100$ & $   0.1000$ & $   0.0200$ & $   0.1267$ & $   0.1067$ & $  -0.0033$ & $   0.0267$ & $   0.1067$ & $   0.0467$ \\
Surround         & $  -9.9833$ & $  -7.8100$ & $  -9.2900$ & $  -9.1433$ & $  -8.9600$ & $  -9.1233$ & $  -9.1100$ & $  -8.9000$ & $  -9.3500$ & $  -8.9767$ \\
\midrule \\[-2.7ex]
IQM(RNS)         & $      0.0$ & $      1.0$ & $   0.3398$ & $   0.4802$ & $   0.4055$ & $   0.4179$ & $   0.3509$ & $   0.4919$ & $   0.3301$ & $   0.3825$ \\
Optimality Gap(RNS) & $      0.0$ & $      1.0$ & $   0.5968$ & $   0.5112$ & $   0.5498$ & $   0.5478$ & $   0.5910$ & $   0.5087$ & $   0.5997$ & $   0.5705$ \\
\bottomrule
\end{tabular}}
\end{center}
\label{table:full_result_suboptimality_far_ood}
\end{table}

\clearpage


\begin{table}[h]
\vspace{-2mm}
\begin{center}
\caption{Offline BC scores on in-distribution games, across models pre-trained with datasets of different sizes.}
\vspace{-2.0ex}
\resizebox{0.75\textwidth}{!}{
\begin{tabular}{r|cc|cccc|cccc}
\multicolumn{3}{c}{} & \multicolumn{4}{c}{\textbf{1M}} &  \multicolumn{4}{c}{\textbf{100M}} \\
\toprule &\\[-2.5ex]
Game & \textit{RndmAgnt} & \textit{DQN} & CURL & ATC & BC & CQL-D & CURL & ATC & BC & CQL-D \\[0.2ex]
\midrule \\[-2.7ex]
AirRaid           & $    579.2$ & $   7479.5$ & $   1025.8$ & $   1398.0$ & $   2772.4$ & $   2799.7$ & $   1499.1$ & $   2232.8$ & $   3613.5$ & $   4359.5$ \\
Amidar            & $      5.8$ & $   1207.7$ & $     14.5$ & $     46.7$ & $     43.8$ & $     45.1$ & $     35.6$ & $     39.5$ & $     77.9$ & $    130.2$ \\
Asteroids         & $    719.1$ & $    698.4$ & $    607.3$ & $    730.3$ & $    817.8$ & $    747.6$ & $    801.9$ & $    732.3$ & $    873.8$ & $    994.1$ \\
Atlantis          & $  12850.0$ & $ 853640.0$ & $  16861.7$ & $  35155.3$ & $   9752.0$ & $  13624.0$ & $  20046.0$ & $  12788.0$ & $  41484.0$ & $  62210.7$ \\
\bottomrule
\end{tabular}}
\end{center}
\label{table:full_result_datascale_id}
\end{table}

\begin{table}[!h]
\vspace{-4.5mm}
\begin{center}
\caption{Offline BC scores on Near-Out-of-Distribution games across models pre-trained with datasets of different sizes.}
\vspace{-2.0ex}
\resizebox{0.72\textwidth}{!}{
\begin{tabular}{r|cc|cccc|cccc}
\multicolumn{3}{c}{} & \multicolumn{4}{c}{\textbf{1M}} &  \multicolumn{4}{c}{\textbf{100M}} \\
\toprule &\\[-2.5ex]
Game & \textit{RndmAgnt} & \textit{DQN} & CURL & ATC & BC & CQL-D & CURL & ATC & BC & CQL-D \\[0.2ex]
\midrule \\[-2.7ex]
Alien          & $    227.8$ & $   2484.5$ & $    463.6$ & $    584.2$ & $    603.9$ & $    500.9$ & $    537.0$ & $    729.0$ & $    639.0$ & $    530.5$ \\
Assault        & $    222.4$ & $   1525.2$ & $    414.0$ & $    663.5$ & $    551.3$ & $    674.7$ & $    575.8$ & $    696.1$ & $    818.8$ & $    635.6$ \\
Asterix        & $    210.0$ & $   2711.4$ & $    275.8$ & $    349.0$ & $    451.8$ & $    412.7$ & $    399.8$ & $    395.8$ & $    550.0$ & $    355.7$ \\
BeamRider      & $    363.9$ & $   5852.4$ & $    542.3$ & $    592.4$ & $    525.4$ & $    468.9$ & $    519.3$ & $    549.3$ & $    572.3$ & $    451.4$ \\
\bottomrule
\end{tabular}}
\end{center}
\label{table:full_result_datascale_near_ood}
\end{table}

\begin{table}[!h]
\vspace{-4.5mm}
\begin{center}
\caption{Offline BC scores on Far-Out-of-Distribution games, across models pre-trained with datasets of different sizes.}
\vspace{-2.0ex}
\resizebox{0.7\textwidth}{!}{
\begin{tabular}{r|cc|cccc|cccc}
\multicolumn{3}{c}{} & \multicolumn{4}{c}{\textbf{1M}} &  \multicolumn{4}{c}{\textbf{100M}} \\
\toprule &\\[-2.5ex]
Game & \textit{RndmAgnt} & \textit{Rainbow} & CURL & ATC & BC & CQL-D & CURL & ATC & BC & CQL-D \\[0.2ex]
\midrule \\[-2.7ex]
BasicMath        & $   0.1334$ & $   3.5700$ & $   1.9300$ & $   3.4233$ & $   2.6400$ & $   2.5633$ & $   3.2167$ & $   3.3333$ & $   2.0967$ & $   3.3600$ \\
HumanCannonball  & $   1.0767$ & $   5.5600$ & $   0.9200$ & $   2.8300$ & $   1.3667$ & $   1.4100$ & $   1.5367$ & $   1.6533$ & $   1.3833$ & $   1.4867$ \\
Klax             & $   0.3734$ & $2124.7500$ & $   5.0167$ & $  16.4667$ & $  23.0667$ & $  12.2667$ & $  12.5667$ & $  23.2167$ & $  51.0667$ & $  23.6833$ \\
Othello          & $ -21.3233$ & $  -2.0100$ & $   0.0000$ & $   0.1667$ & $  -0.1000$ & $  -0.0533$ & $  -0.2800$ & $   0.3867$ & $  -0.0800$ & $   0.1867$ \\
Surround         & $  -9.9833$ & $  -7.8100$ & $  -9.4967$ & $  -9.1200$ & $  -8.9233$ & $  -9.1800$ & $  -9.1967$ & $  -8.9833$ & $  -8.8833$ & $  -8.9367$ \\
\midrule \\[-2.7ex]
IQM(RNS)         & $      0.0$ & $      1.0$ & $   0.2498$ & $   0.5819$ & $   0.4273$ & $   0.3837$ & $   0.4539$ & $   0.5093$ & $   0.3819$ & $   0.5040$ \\
Optimality Gap(RNS) & $      0.0$ & $      1.0$ & $   0.6572$ & $   0.4494$ & $   0.5415$ & $   0.5687$ & $   0.5265$ & $   0.4958$ & $   0.5661$ & $   0.4981$ \\
\bottomrule
\end{tabular}}
\end{center}
\label{table:full_result_datascale_far_ood}
\end{table}


\begin{table}[h]
\vspace{-2mm}
\begin{center}
\caption{Offline BC scores of a smaller backbone(ResNet-18) on In-Distribution games.}
\vspace{-2.0ex}
\resizebox{0.65\textwidth}{!}{
\begin{tabular}{r|cc|cccccc}
\multicolumn{3}{c}{} & \multicolumn{6}{c}{\textbf{ResNet-18}} \\
\toprule &\\[-2.5ex]
Game & \textit{RndmAgnt} & \textit{DQN} & Random & E2E & CURL & ATC & BC & CQL-D \\[0.2ex]
\midrule \\[-2.7ex]
AirRaid           & $    579.2$ & $   7479.5$ & $   1550.9$ & $   4671.4$ & $   1100.7$ & $   1305.3$ & $   3180.2$ & $   3306.7$ \\
Amidar            & $      5.8$ & $   1207.7$ & $     21.3$ & $     70.5$ & $     18.8$ & $     34.1$ & $     68.3$ & $     56.1$ \\
Asteroids         & $    719.1$ & $    698.4$ & $    872.3$ & $    792.7$ & $    612.5$ & $    670.8$ & $    777.7$ & $    731.1$ \\
Atlantis          & $  12850.0$ & $ 853640.0$ & $  10249.0$ & $ 133756.0$ & $  16177.0$ & $  17939.3$ & $  12289.7$ & $  16609.7$ \\
\bottomrule
\end{tabular}}
\end{center}
\label{table:full_result_modelscale_id}
\end{table}

\begin{table}[!h]
\vspace{-1.5mm}
\begin{center}
\caption{Offline BC scores of a smaller backbone(ResNet-18) on Near-Out-of-Distribution games.}
\vspace{-2.0ex}
\resizebox{0.65\textwidth}{!}{
\begin{tabular}{r|cc|cccccc}
\multicolumn{3}{c}{} & \multicolumn{6}{c}{\textbf{ResNet-18}} \\
\toprule &\\[-2.5ex]
Game & \textit{RndmAgnt} & \textit{DQN} & Random & E2E & CURL & ATC & BC & CQL-D \\[0.2ex]
\midrule \\[-2.7ex]
Alien          & $    227.8$ & $   2484.5$ & $    542.8$ & $    618.0$ & $    402.0$ & $    739.7$ & $    777.9$ & $    711.6$ \\
Assault        & $    222.4$ & $   1525.2$ & $    405.7$ & $   1004.2$ & $    410.1$ & $    641.8$ & $    677.4$ & $    668.7$ \\
Asterix        & $    210.0$ & $   2711.4$ & $    352.3$ & $    436.2$ & $    293.3$ & $    414.8$ & $    443.2$ & $    406.7$ \\
BeamRider      & $    363.9$ & $   5852.4$ & $    594.5$ & $    730.8$ & $    542.7$ & $    487.9$ & $    468.0$ & $    467.1$ \\
\bottomrule
\end{tabular}}
\end{center}
\label{table:full_result_modelscale_near_ood}
\end{table}

\begin{table}[!h]
\vspace{-1.5mm}
\begin{center}
\caption{Offline BC scores of a smaller backbone(ResNet-18) on Far-Out-of-Distribution games.}
\vspace{-2.0ex}
\resizebox{0.58\textwidth}{!}{
\begin{tabular}{r|cc|ccccc}
\multicolumn{3}{c}{} & \multicolumn{5}{c}{\textbf{ResNet-18}} \\
\toprule &\\[-2.5ex]
Game & \textit{RndmAgnt} & \textit{Rainbow} & Random & CURL & ATC & BC & CQL-D \\[0.2ex]
\midrule \\[-2.7ex]
BasicMath        & $   0.1334$ & $   3.5700$ & $   1.1500$ & $   2.7667$ & $   3.6667$ & $   2.9133$ & $   3.3833$ \\
HumanCannonball  & $   1.0767$ & $   5.5600$ & $   0.9900$ & $   1.1367$ & $   2.6800$ & $   1.1400$ & $   1.6867$ \\
Klax             & $   0.3734$ & $2124.7500$ & $   3.6667$ & $   6.9833$ & $  27.1667$ & $  20.0667$ & $  23.3167$ \\
Othello          & $ -21.3233$ & $  -2.0100$ & $   0.0000$ & $   0.5667$ & $   0.0333$ & $   0.0600$ & $   0.2733$ \\
Surround         & $  -9.9833$ & $  -7.8100$ & $  -9.8967$ & $  -9.0700$ & $  -9.0000$ & $  -8.7333$ & $  -8.8700$ \\
\midrule \\[-2.7ex]
IQM(RNS)         & $      0.0$ & $      1.0$ & $   0.1134$ & $   0.4029$ & $   0.6110$ & $   0.4790$ & $   0.5313$ \\
Optimality Gap(RNS) & $      0.0$ & $      1.0$ & $   0.7364$ & $   0.5594$ & $   0.4409$ & $   0.5185$ & $   0.4790$ \\
\bottomrule
\end{tabular}}
\end{center}
\label{table:full_result_modelscale_far_ood}
\end{table}

\end{document}